\documentclass{amsart}

\usepackage{microtype}
\usepackage{graphicx}
\usepackage{subfigure}
\usepackage{booktabs} 

\usepackage{hyperref}

\usepackage{commands/nkj}
\usepackage[utf8]{inputenc}

\usepackage{amsmath}
\usepackage{amsfonts}
\usepackage{xcolor}
\usepackage{graphicx}

\usepackage{multirow}
\usepackage{multicol}
\usepackage{todonotes}
\usepackage{tabularx}

\usepackage{tikz}
\usetikzlibrary{positioning}
\usetikzlibrary{shapes,arrows}
\usepackage{tikz, pgfplots}
\pgfplotsset{compat=1.14}
\newlength{\figureheight}
\newlength{\figurewidth}
\tikzstyle{line} = [draw, -latex']
\tikzstyle{visible_box} = [regular polygon,regular polygon sides=4, draw,rounded corners, minimum size = 1.2cm]
\tikzstyle{invisible_box} = [regular polygon,regular polygon sides=4,rounded corners, minimum size = 1.2cm]

\tikzstyle{input} = [rectangle, draw, 
    text width=1em, rounded corners, minimum height=6em]
\tikzstyle{text_processor} = [rectangle, draw, 
    text width=7em, rounded corners, minimum height=6em]
\tikzstyle{note_box} = [rectangle, draw, dashed,
    text width=8em, rounded corners, minimum height=10em]
\tikzstyle{note_box2} = [rectangle, draw, dashed,
    text width=16em, rounded corners, minimum height=10em]
\tikzstyle{invisible_box} = [rectangle, text width=13em, rounded corners, minimum height=8em]
\tikzstyle{invisible_box2} = [rectangle, text width=20em, rounded corners, minimum height=8em]
\tikzstyle{class_box} = [rectangle, draw, text width = 2em, rounded corners, minimum height=2em]
\tikzstyle{line} = [draw, -latex']

\newcommand{\DropLong}{Monte Carlo dropout}

\newcommand{\IntervalLong}{Interval Neural Network}
\newcommand{\DropShort}{\textsc{MCDrop}}
\newcommand{\ProbShort}{\textsc{ProbOut}}
\newcommand{\IntervalShort}{\textsc{INN}}

\newcommand{\DeconvNNShort}{DeconvNet}

\newcommand{\DeconvLong}{1D Deconvolution}

\newcommand{\CTLong}{Computed Tomography Reconstruction}
\newcommand{\DeconvShort}{\textsc{1DDeconv}}

\newcommand{\CTShort}{\textsc{CT}}

\newcommand{\ErrorLong}{Error Correlation}

\newcommand{\ErrorShort}{EC}

\usepackage{amsthm, amssymb}
\usepackage{cleveref}
\usepackage[foot]{amsaddr}

\input{dump/abbrv}

\usepackage{appendix}
\crefname{appsec}{Appendix}{Appendices}

\begin{document}

\title{Interval Neural Networks: Uncertainty Scores}

\author[L. Oala]{Luis Oala\textsuperscript{$\star$1}}
\address{\textsuperscript{1}Machine Learning Group, Fraunhofer HHI,
10587 Berlin, Germany}
\email{\{luis.oala,wojciech.samek\}@hhi.fraunhofer.de}
\thanks{\textsuperscript{$\star$}Equal contribution}

\author[C. Heiß]{Cosmas Heiß\textsuperscript{$\star$2}} 
\address{\textsuperscript{2}Department of Mathematics, Technical University of Berlin, 10623 Berlin, Germany}
\email{cosmas.heiss@gmail.com}
\thanks{The authors would like to thank Sören Becker for feedback on the final draft. J.M. acknowledges support by DFG-RTG 2260 BIOQIC. M.M. acknowledges support by the DFG Priority Programme DFG-SPP 1798 Grants KU 1446/21 and KU 1446/23. G.K. is grateful to MATH+ -- Berlin Mathematics Research Center (Project EF1x1) for financial support.}

\author[J. Macdonald]{Jan Macdonald\textsuperscript{2}}
\email{\{macdonald,maerz,kutyniok\}@math.tu-berlin.de}
\author[M. März]{Maximilian März\textsuperscript{2}}
\author[W. Samek]{Wojciech Samek\textsuperscript{1}}
\author[G. Kutyniok]{Gitta Kutyniok\textsuperscript{2}}

\keywords{Deep Learning, Uncertainty Quantification, Inverse Problems}

\begin{abstract}
We propose a fast, non-Bayesian method for producing uncertainty scores in the output of pre-trained deep neural networks (DNNs) using a data-driven interval propagating network. This interval neural network (\IntervalShort) has interval valued parameters and propagates its input using interval arithmetic.
The \IntervalShort\ produces sensible lower and upper bounds encompassing the ground truth. We provide theoretical justification for the validity of these bounds. Furthermore, its asymmetric uncertainty scores offer additional, directional information beyond what Gaussian-based, symmetric variance estimation can provide. We find that noise in the data is adequately captured by the intervals produced with our method.
In numerical experiments on an image reconstruction task, we demonstrate the practical utility of \IntervalShort s as a proxy for the prediction error in comparison to two state-of-the-art uncertainty quantification methods.
In summary, \IntervalShort s produce fast, theoretically justified uncertainty scores for DNNs that are easy to interpret, come with added information and pose as improved error proxies - features that may prove useful in advancing the usability of DNNs especially in sensitive applications such as health care.
\end{abstract}

\maketitle
\section{Introduction}
Deep neural networks (DNNs) nowadays play a remarkable role in many computational imaging tasks such as image translation (domain mapping), super-resolution, denoising or image synthesis. 
Specifically, inverse problems in medical imaging, a class of problems crucial for technologies such as computed tomography (CT) or magnetic resonance imaging, are a promising field of application for DNNs, see \cite{arridge_maass_oektem_schoenlieb_2019} for a recent overview. Optimization and further automation in this setting can have positive impacts to increase the quality and coverage of medical care.
So far, the computational prowess of DNNs comes at the cost of opacity, a situation which extinguishes their prospects of adoption in a medical setting. A growing body of work has accumulated over recent years aimed at mitigating this opacity. An important strand concerns so called \textit{uncertainty quantification} (UQ), an umbrella term for methods that provide confidence scores for predictions of a variety of models including DNNs.
\par
Existing UQ methods come with limitations regarding error correlations, a lack of directional expressiveness due to uncertainty symmetry and non-robust behavior in the face of noise. We present a novel, non-Bayesian and optimization-based UQ method called the \textit{\IntervalLong} \ (\IntervalShort), which provides improvements in alleviating these limitations. An \IntervalShort\ is an interval valued neural network that is trained to encompass the target values while being bound to the behavior of an underlying prediction network. As we will demonstrate, this leads to improved error correlation as well as uncertainty scores with increased expressiveness. Our method also features improved robustness to input noise and analytic coverage bounds while maintaining competitive run-time. Finally, we perform experiments on an image reconstruction task to corroborate the capacities of the \IntervalShort\ method empirically vis-a-vis two existing, popular UQ methods for DNNs.
\section{Related Work}
\label{sec:related-work}
Whereas a number of methods from classical statistical learning theory, such as Gaussian processes and approximations thereof \cite{denker_large_1987, mackay_bayesian_1992, hinton_bayesian_1995, williams_computing_1996}, come with built-in uncertainty estimates, DNNs have been limited in this regard. A surge of efforts to treat neural networks from a variational perspective \cite{barber_ensemble_nodate, srivastava_dropout:_2014, blundell_weight_2015, kingma_variational_2015} started to change that and led to \cite{gal_dropout_2016,kendall_what_2017}’s now widespread  \DropLong\ (\DropShort) framework which we will also use in all experiments as a baseline. In \DropShort\ mean and variance estimates for a DNN prediction are obtained by calculating the sample mean and variance on multiple stochastic forward passes on the same input data point. The stochasticity is induced by performing dropout during inference and sampling a new realization from the dropout Bernoulli distribution at each pass. Earlier, \cite{nix_estimating_1994} had already proposed another simple recipe for uncertainty estimates which are amicable to the DNN treatment: the number of output components of a DNN is doubled and the DNN is trained to approximate the mean and variance of a Gaussian distribution. This led to the introduction of lightweight probabilistic DNNs by \cite{gast_lightweight_2018}, dubbed \ProbShort, which we will use as the second baseline. Alternate, less popular approaches to UQ for neural networks are based on confidence intervals or slight variations thereof. \cite{khosravi_comprehensive_2011} provides an overview for the interested reader. Most notably, \cite{shafer_tutorial_2008} developed what they call \textit{conformal prediction}. Its application is limited, though, since an exchangeability assumption on samples is made which rarely holds in a practical setting.

Despite the fact that deep learning based methods are becoming state-of-the-art for solving various inverse problems, this field is still in an early stage. Most approaches in this direction focus on the empirical reconstruction performance, excluding other aspects such as uncertainty quantification. A notable exception is a recent work by \cite{adler_deep_2018}, who made use of a Bayesian framework to also consider uncertainties in their reconstruction.
\section{Research Needs}
Existing UQ methods for DNNs provide confidence estimates. Low uncertainty does not necessarily imply a low prediction error. In fact, a confident prediction may be right or wrong. Furthermore, the uncertainty scores of existing UQ methods are symmetric, e.g. obtained from Gaussian predictive densities. Thus, if a practitioner sees high uncertainty scores in a reconstructed medical image, she cannot infer the direction of uncertainty: is the DNN uncertain whether the output values could rather be higher or lower? 
Finally, as we demonstrate in our experiments, scores by existing UQ methods can become detached from the behavior of the underlying prediction model in the face of perturbations on the data.
\par
With our \IntervalShort\ method we aim to alleviate these limitations and increase the utility of uncertainty scores. We achieve this by propagating intervals through the network yielding a high-dimensional box in the output space, which necessarily contains the prediction and is fitted to also encompass the target data. The \IntervalShort\ uncertainty scores thus provide information about the prediction error, the direction of uncertainty and are more robust to noise perturbations in the input and output data than existing UQ methods. Finally, the \IntervalShort\ method comes with the marked benefit of analytic coverage bounds that allow us to specify the proportion of targets that fall outside the predicted intervals. All of these benefits can be obtained at competitive or improved run-time vis-a-vis the existing baseline methods.

On the applied side this work is motivated by the application of deep learning for solving inverse problems in medical imaging. The most common type are linear inverse problems, which can be written as
\begin{equation}\label{eq:inv_prob}
 \bfx = \bfA \bfy + \bfeta,
\end{equation}
\begin{figure}[t]
    \centering
    \input{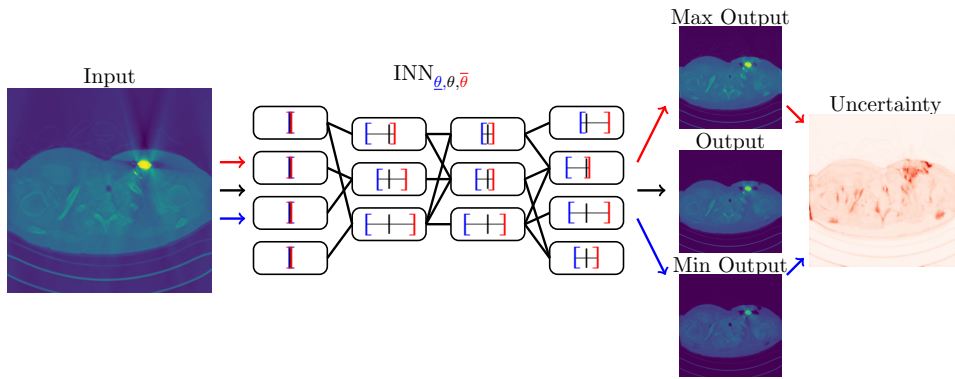}
    \caption{\textbf{\IntervalShort\ Schematic Overview.} The structure of an \IntervalLong\ and how it is used to compute uncertainty scores. The input (left picture) is interpreted as a point interval in the first layer. It is then propagated through the network by the interval valued weights and biases (black connections) using interval arithmetic. As the weights are constrained using the underlying network, the interval valued neurons contain the value from the original prediction (black bar inside intervals) in every layer. In the output, three images are obtained: the original prediction and the images containing the lower and upper bounds. The latter two can then be used to construct a pixel-wise uncertainty score from the interval size.}
    \label{fig:interval_overview}
\end{figure}where $\bfy\in\R^n$ is the unknown signal of interest, $\bfA\in\R^{m\times n}$ is the forward operator representing a physical measurement process, and $\bfeta\in\R^m$ is modelling noise in the measurements. Typical examples include choosing $\bfA$ as the identity (denoising), a subsampled Fourier matrix (magnetic resonance imaging), or a discrete Radon transform (computed tomography).  
Solving the inverse problem \eqref{eq:inv_prob} amounts to reconstructing $\bfy$ from its observed measurements $\bfx$. Recently, data driven methods (in particular deep learning based reconstructions, trained on given sample pairs $(\bfx_i, \bfy_i)$) have shown the potential to outperform earlier model based regularization schemes \cite{arridge_maass_oektem_schoenlieb_2019}. The risks of employing deep learning for reconstruction are their lack of theoretical recovery guarantees and the susceptibility to unwanted artifacts \cite{vegard2019,huang2018}. In particular, for sensitive medical imaging applications, it is indispensable to obtain more information regarding the reliability of the reconstruction method in the form of uncertainty quantification.
\section{Interval Neural Networks}
\label{seq:method}
Popular existing UQ frameworks for DNNs rely on placing parametric densities, most commonly a Gaussian density, over DNN parameters or predictions. Our INN method relies on bounding this distribution using intervals. This is based upon the idea of \cite{garczarczyk_interval_2000} for investigating the capacity of neural networks with interval \textit{weights} and \textit{biases} for fitting interval valued functions. Note that \cite{kowalski_interval_2017} also explored the usage of interval neural networks for robust classification although in their setting the focus is purely on representing the \textit{inputs} as intervals. Our resulting INN is therefore to some extent similar to existing interval propagation models except that interval bounds are determined for all parameters of the network with the goal of providing uncertainty scores for the outputs.
\par
For an existing DNN to be analyzed, which we will from now on call the \textit{underlying model}, the \IntervalShort\ method consists of an interval valued neural network which has the same architecture as the underlying, existing DNN. The \IntervalShort\ is initialized with the parameters of the underlying model as point intervals. It then operates as follows: Any input vector is first regarded as interval valued with point interval components. This interval vector is then propagated through the network. As illustrated in \Cref{fig:interval_overview} the interval sizes in the resulting output are regarded as uncertainty estimates. The interval parameters are trained on the training data using the loss function in \Cref{eq:interval-loss}, which we will explain in more detail below. The goal of the training procedure for the \IntervalShort\ is to produce output intervals that contain the true labels with high probability, while remaining as tight as possible. The \IntervalShort 's intervals are constrained to always contain the parameters of the underlying model. This constraint not only helps with stability during training but also guarantees that the output of the underlying network is always contained in the output interval of the \IntervalShort. Thus, output interval sizes serve the purpose of upper bounding the absolute error.
\par
\IntervalShort s have the following mechanisms that deviate from the customary arithmetic. The forward propagation of a component-wise interval valued input $[\underline{\bfx}, \overline{\bfx}]$ through the INN can be expressed similarly to standard feed-forward neural networks but using interval arithmetic instead. For interval valued weight matrices $\left[\underline{\bfW}, \overline{\bfW}\right]$ and bias vectors $\left[\underline{\bfb}, \overline{\bfb}\right]$ the propagation through the $l$-th layer can be expressed using interval arithmetic by
\begin{equation*}
\begin{aligned}
    \left[\underline{\bfx}, \overline{\bfx}\right]^{(l+1)} = \varrho
    \left(\left[\underline{\bfW}, \overline{\bfW}\right]^{(l)}
    \left[\underline{\bfx}, \overline{\bfx}\right]^{(l)}+
    \left[\underline{\bfb}, \overline{\bfb}\right]^{(l)} \right).
\end{aligned}\label{eq:interval_prop}
\end{equation*}
For positive values of $[\underline{\bfx}, \overline{\bfx}]^{(l)}$, for example when using a non-negative activation function like ReLU, we can simplify this operation to
\begin{equation*}
\medmuskip=-1.mu
\thinmuskip=0.mu
\thickmuskip=-1mu
\begin{aligned}
    \overline{\bfx}^{(l+1)} &= \varrho \left(
    \min \left\{
    \overline{\bfW}^{(l)},0\right\}
    \underline{\bfx}^{(l)}
     +\max \left\{
    \overline{\bfW}^{(l)}, 0 \right\}
    \overline{\bfx}^{(l)}+
    \overline{\bfb}^{(l)}
    \right),\\
    \underline{\bfx}^{(l+1)} &= \varrho \left(
    \max \left\{
    \underline{\bfW}^{(l)}, 0\right\}
    \underline{\bfx}^{(l)}+
    \min \left\{
    \underline{\bfW}^{(l)}, 0\right\}
    \overline{\bfx}^{(l)}+
    \underline{\bfb}^{(l)}        
    \right),
\end{aligned}
\end{equation*}
where the maximum and minimum functions are applied component-wise. Assuming $\underline{\bfx}^{(l)}=\overline{\bfx}^{(l)}=:\bfx^{(l)}$ for the input layer before the first ReLU, i.e.\ if the components consist of point intervals, the same operation can be represented as follows to also process negative values:
\begin{equation*}
\medmuskip=0.mu
\thinmuskip=0.mu
\thickmuskip=0.mu
\begin{aligned}
    \overline{\bfx}^{(l+1)} &= \varrho \left(
    \overline{\bfW}^{(l)}
    \max \{
    \bfx^{(l)},0 \}+
    \underline{\bfW}^{(l)}
    \min \{
    \bfx^{(l)},0 \}+
    \overline{\bfb}^{(l)}
    \right),\\
    \underline{\bfx}^{(l+1)} &= \varrho \left(
    \underline{\bfW}^{(l)}
    \max \{
    \bfx^{(l)},0 \}+
    \overline{\bfW}^{(l)}
    \min \{
    \bfx^{(l)},0 \}+
    \underline{\bfb}^{(l)}        
    \right).
\end{aligned}
\end{equation*}
These formulas can then easily be used in existing deep learning frameworks to optimize the bounds of the interval parameters by means of backpropagation.
As we want the output intervals to contain the target values after training, we define the interval loss to be zero if a target lies inside the interval and the squared distance to the interval boundary if it lies outside the interval. As this alone would lead to the intervals in the output expanding until they cover the whole range of target values, we additionally employ a linear penalty on the interval size. For the data set $\left\{\bfx_i,\bfy_i \right\}_{i=1}^m$ consisting of inputs $\bfx_i \in \CX$ and targets $\bfy_i \in \CY$, this leads to the following \IntervalShort\ loss. Here, $\overline{\bfPhi}\colon \CX \to \CY$, $\underline{\bfPhi}\colon \CX \to \CY$ are the functions that map the input to the upper and the lower interval bounds in the output of the \IntervalShort :
\begin{align}
\medmuskip=0.mu
\thinmuskip=0.mu
\thickmuskip=0.mu
    \mathcal{L}(\underline{\bfPhi},\overline{\bfPhi}) =&
    \sum_{i=1}^{m}\max \{\bfy_i-\overline{\bfPhi}(\bfx_i),0\}^2\nonumber\\
    &\quad+\max\{\underline{\bfPhi}(\bfx_i)-\bfy_i,0\}^2
    +\beta \cdot \big(\overline{\bfPhi}(\bfx_i)-\underline{\bfPhi}(\bfx_i) \big).\label{eq:interval-loss}
\end{align}
The tightness parameter $\beta > 0$ determines how outlier-sensitive the intervals are trained. In practice, choosing $\beta$ similar to the mean absolute error made by the underlying prediction network seems to be a good heuristic. 
\section{Qualities of Interval Neural Networks}
\label{sec:strengths}
\IntervalShort s come with certain features that go beyond what other methods for UQ can provide. These features comprise theoretically justified coverage bounds, adaptive behavior under noise as well as directional information in the uncertainty score. 
\par
\begin{figure*}[ht]
    \centering
    \begin{tabular}{rrr}
        &\includegraphics[width=0.425\textwidth]{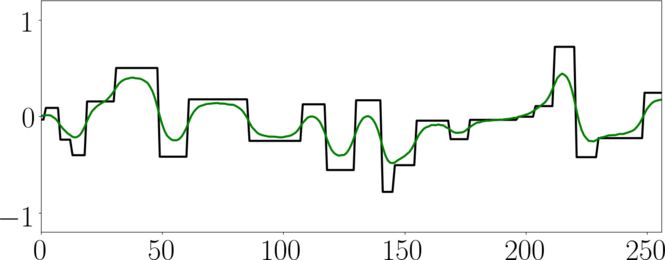}&
        \includegraphics[width=0.425\textwidth]{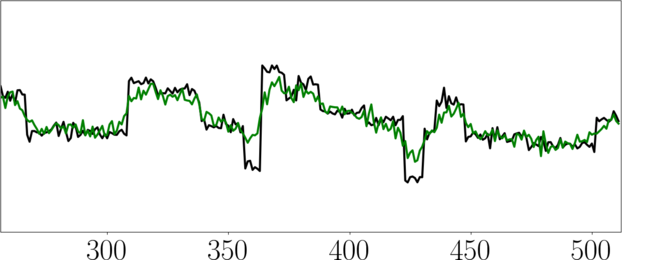} \\
        
        \multirow[t]{2}{*}{(I)$\begin{cases}$\vspace*{3.3cm}$\end{cases}$ \hspace*{-0.8cm}}
        &\includegraphics[width=0.425\textwidth]{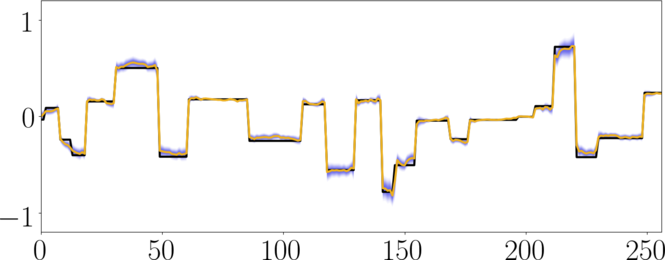}& 
        \includegraphics[width=0.425\textwidth]{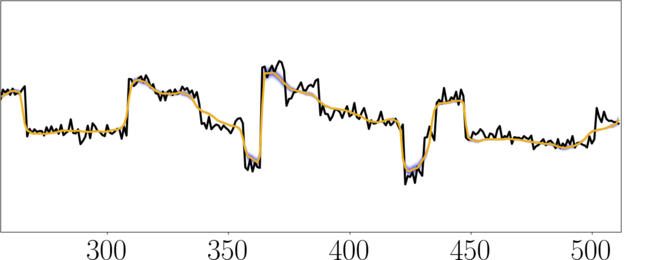} \\
        
        &\includegraphics[width=0.425\textwidth]{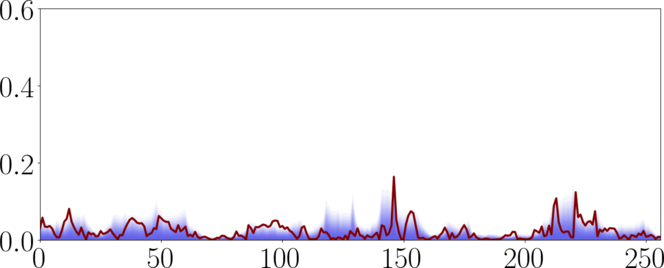}& 
        \includegraphics[width=0.425\textwidth]{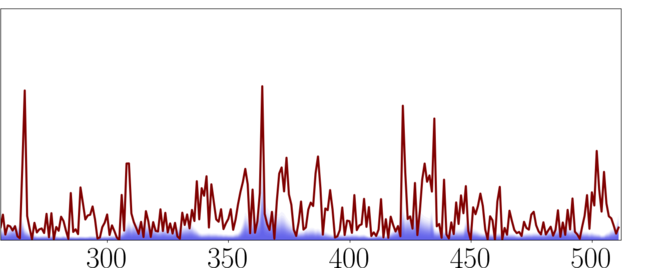} \\
        
        \multirow[t]{2}{*}{(II)$\begin{cases}$\vspace*{3.3cm}$\end{cases}$ \hspace*{-0.8cm}}
        &\includegraphics[width=0.425\textwidth]{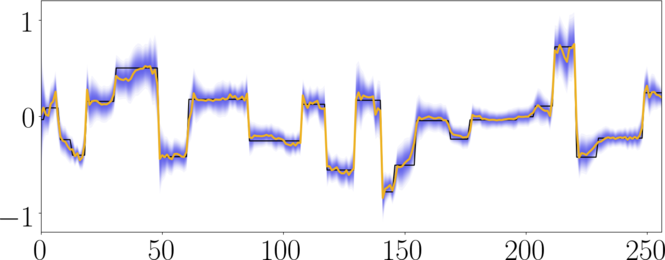}& 
        \includegraphics[width=0.425\textwidth]{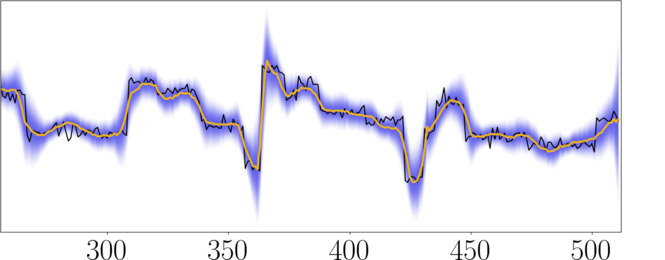} \\
        
        &\includegraphics[width=0.425\textwidth]{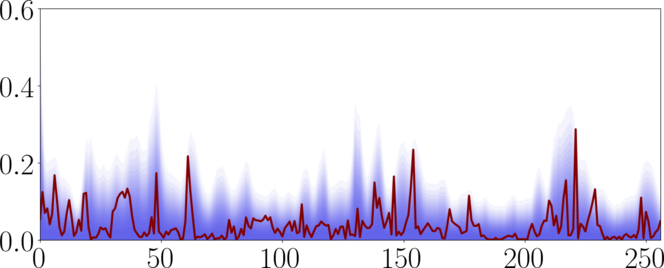}&
        \includegraphics[width=0.425\textwidth]{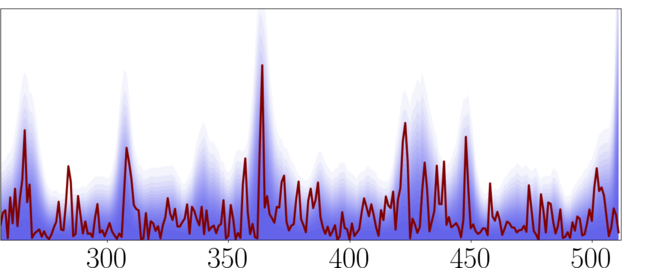} \\
        
        \multirow[t]{2}{*}{(III)$\begin{cases}$\vspace*{3.3cm}$\end{cases}$ \hspace*{-0.8cm}}
        &\includegraphics[width=0.425\textwidth]{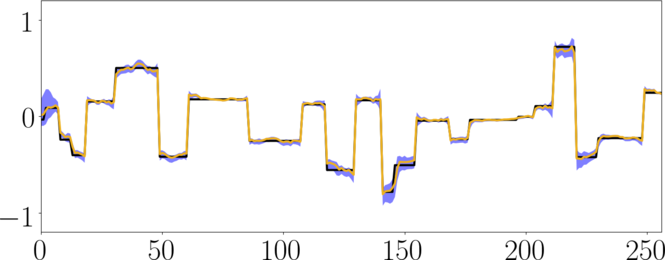}&
        \includegraphics[width=0.425\textwidth]{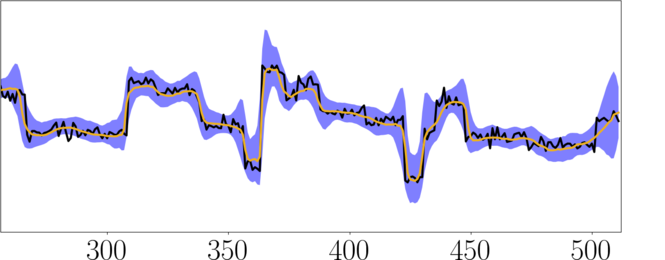} \\
        &\includegraphics[width=0.425\textwidth]{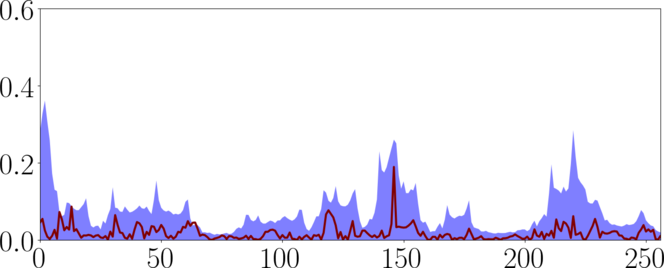}& 
        \includegraphics[width=0.425\textwidth]{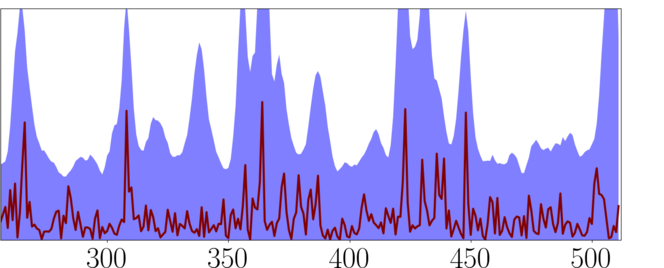} \\
        
        &\multicolumn{2}{l}{\hspace{0.026\textwidth}\includegraphics[width=0.77\textwidth]{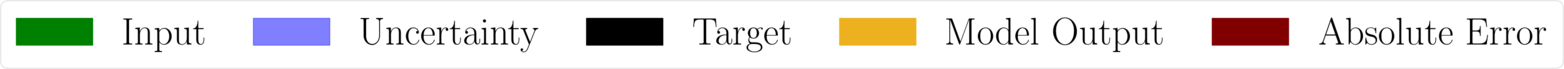}}
    \end{tabular}
    \caption{\textbf{\textsc{1DDeconv} Task Results.} Results for the \textsc{1DDeconv} task on the same sample without noise on the left and with Gaussian noise ($\sigma=0.05$) on the right. The first row displays input and target vector. The figures below show target, network output, together with the uncertainty estimation for the upper graph and the uncertainty estimation plotted against the absolute error in the lower graph for each corresponding method; (I) \textsc{MCDrop}, (II) \textsc{ProbOut}, (III) \textsc{INN}.}
    \label{fig:1d_example_images1}
\end{figure*}

In order to demonstrate these qualities, we devised an illustrative toy task as follows. Note that we also provide additional, quantitative experiments on real-life CT data in \Cref{sec:Experiments}.
The toy experiment (\DeconvShort) is based on a simple use case of an ill-conditioned inverse problem, which is inspired by a one-dimensional deconvolution task. We choose $\bfA = \bfD^\top \bfS \bfD\in\R^{512\times 512}$, where $\bfD\in\R^{512\times 512}$ is a discrete cosine transform and $\bfS\in\R^{512\times512}$ is a diagonal matrix with exponentially decaying values. We consider discretizations of piecewise constant functions with random jump positions and heights as the signal distribution in $\R^{512}$. The blurred measurements $\bfx\in\R^{512}$ corresponding to each signal sample $\bfy$ are simulated by computing $\bfx=\bfA \bfy+\bfeta$ as in \eqref{eq:inv_prob} (see top of \Cref{fig:1d_example_images1} for an illustration).
The considered data set consists of $2000$ sample pairs $(\bfx_i, \bfy_i)$, $1600$ of which were used for training, $200$ for validation and $200$ for testing. This one-dimensional data allows 
\noindent for good illustration of the different INN characteristics. The prediction DNN for the \DeconvShort\ task, called \DeconvNNShort, consists of a convolutional neural network (CNN) trained to directly map $\bfx$ to $\bfy$. It consists of 10 convolutional layers and three dropout layers, one with dropout probability $0.2$ and the other two with probability $0.5$. The number of channels increases through the first $7$ layers to $256$ and decreases back to $1$ in the successive layers. No pooling is employed and the data size of each channel is held to be the same as the input size throughout the network. The prediction DNN was trained for $100$ epochs using Adam \cite{kingma2014adam} with a learning rate of $10^{-3}$ and batch size $256$. The interval parameters of the \IntervalShort\ were then trained for another $100$ epochs with a learning rate of $10^{-5}$ and $\beta=2\times 10^{-3}$. For the \DropShort\ comparison, 64 samples were used to estimate mean and variance and for the \ProbShort\ comparison, the \ProbShort\ loss was also optimized for 100 epochs using the Adam optimizer with a learning rate of $10^{-4}$.
\par Let us now turn to the illustration of the \IntervalShort 's qualities in this setting. First, the \IntervalShort\ loss is directly tied to the target data, which enables us to bound the portion of the targets that are contained in the intervals of a given model and data set. This is not possible with the popular Gaussian-based UQ methods. In the case of the \DeconvShort\ task the portion of ground truth values that lie inside the intervals is $89\%$. What is more, these coverage bounds can be theoretically justified. They are an innate quality of the \IntervalShort\ approach that follows from its loss function. 
Assuming the loss in \Cref{eq:interval-loss} is optimized during training to yield an \IntervalShort\ for which the expected gradient with respect to the data distribution is zero, we can make the following estimates using the Markov Inequality: Let the training data be represented by the random variable $(\bfx^*,\bfy^*)$ being distributed on $\CX \times \CY$ according to the training data distribution. Then, for any $\lambda > 0$, we obtain
\begin{equation*}
\begin{aligned}
    \mathbb{P}(\underline{\bfPhi}(\bfx^*)-\lambda \beta < \bfy^* < \overline{\bfPhi}(\bfx^*)+\lambda \beta \ | \bfx^*) \geq 1-\frac{1}{\lambda},
\end{aligned}\label{eq:cov-bound-1}
\end{equation*}
i.e.\ for any input and target sampled form the distribution of training samples, the probability of the target lying inside the predicted interval, which is enlarged by $\lambda \beta$, is at least $1-\frac{1}{\lambda}$. As $\beta$ is usually very small, this ensures a fast decay of the probability of a target value lying beyond the interval bounds. Of course, this estimate only holds for the training distribution so the training set has to be chosen to sufficiently represent the true data distribution. An example of the interval consistently containing the target values can be observed in \Cref{fig:1d_example_images1}.
\par
\begin{figure}[t]
    \centering
    \includegraphics[width=0.6\textwidth]{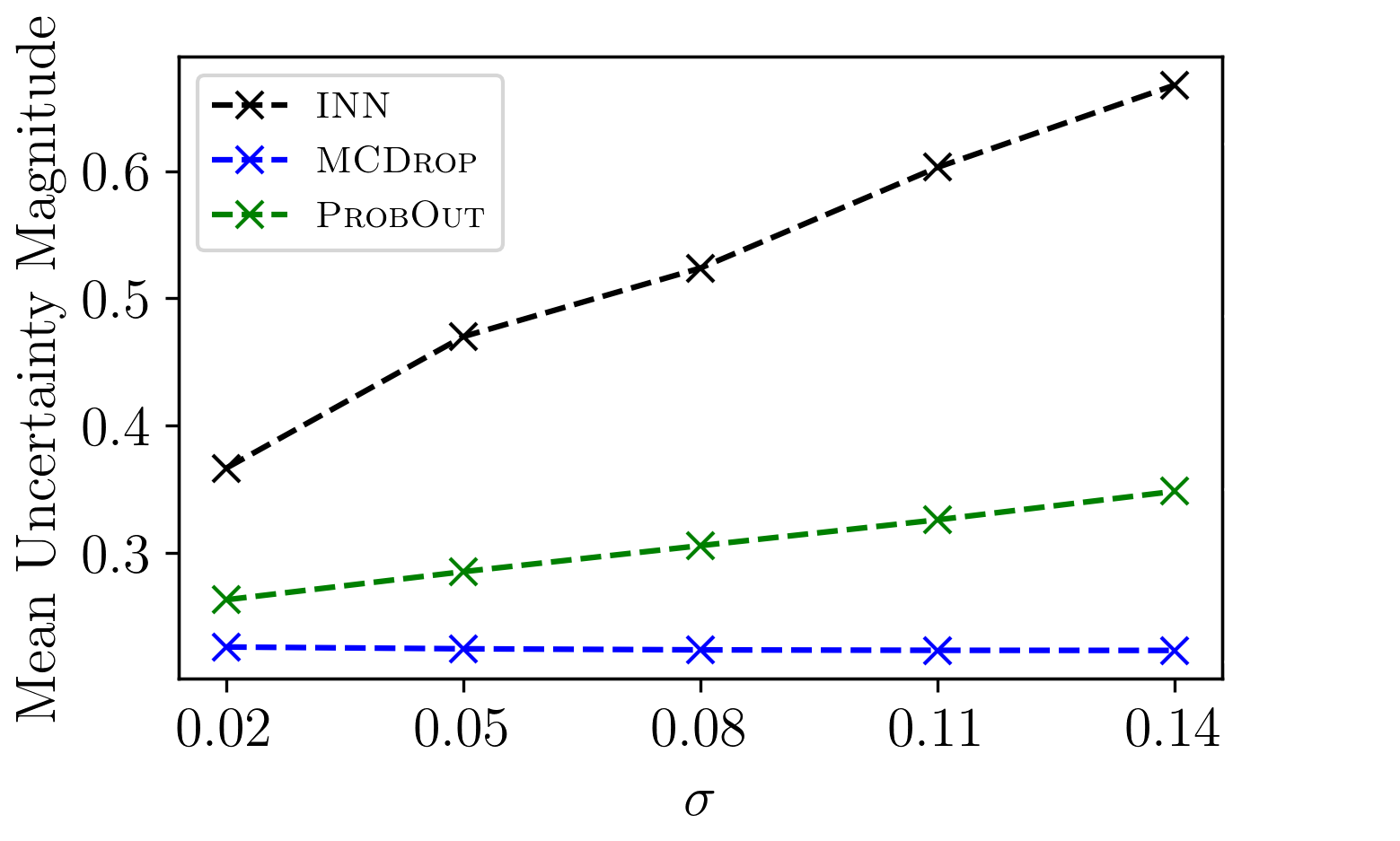}
    \caption{\textbf{Noise Behavior.} Relationship of the mean uncertainty magnitude and additive Gaussian noise on the \DeconvShort\ task. The standard deviation of the additive Gaussian noise for the input and target data is displayed on the x-axis. The mean uncertainty magnitude, which is measured in interval size for \IntervalShort\ (in black) and standard deviation for \DropShort\ (in blue) and \ProbShort\ (in green) meaned over the test data, is displayed on the y-axis.}
    \label{fig:noise_comparison_1D}
\end{figure}

 Second, in many tasks, unpredictable variables, e.g.\ noise in the data, pose an upper bound to the performance of a model. Due to the \IntervalShort\ loss function, the output intervals will be able to capture this noise using the interval bias parameters in the last layer. Therefore, our method will indicate uncertainty even if the underlying network predicts the mean in the data very consistently. This effect is visible in the right column of \Cref{fig:1d_example_images1} for the \DeconvShort\ task with independent Gaussian noise ($\sigma=0.05$) added to the inputs and targets. Note how the \DropShort\ approach is not able to capture these deviations in the output as the network is trained to steadily predict the mean. Furthermore, in \Cref{fig:noise_comparison_1D} one can follow how the average interval size increases with increasing noise levels.
\par
\begin{figure}[t]
    \centering
    \includegraphics[width=0.6\textwidth]{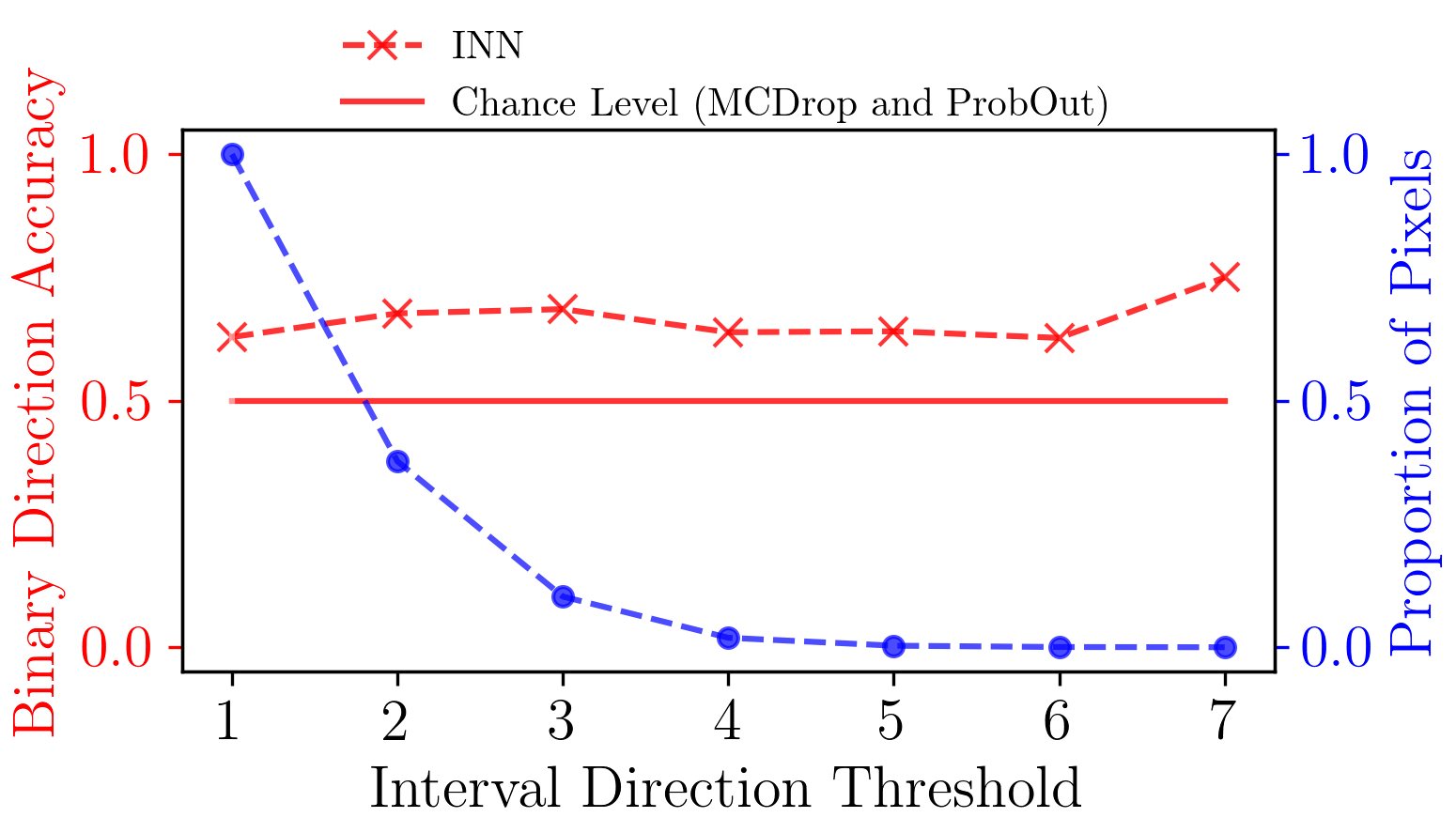}
    \includegraphics[width=0.6\textwidth]{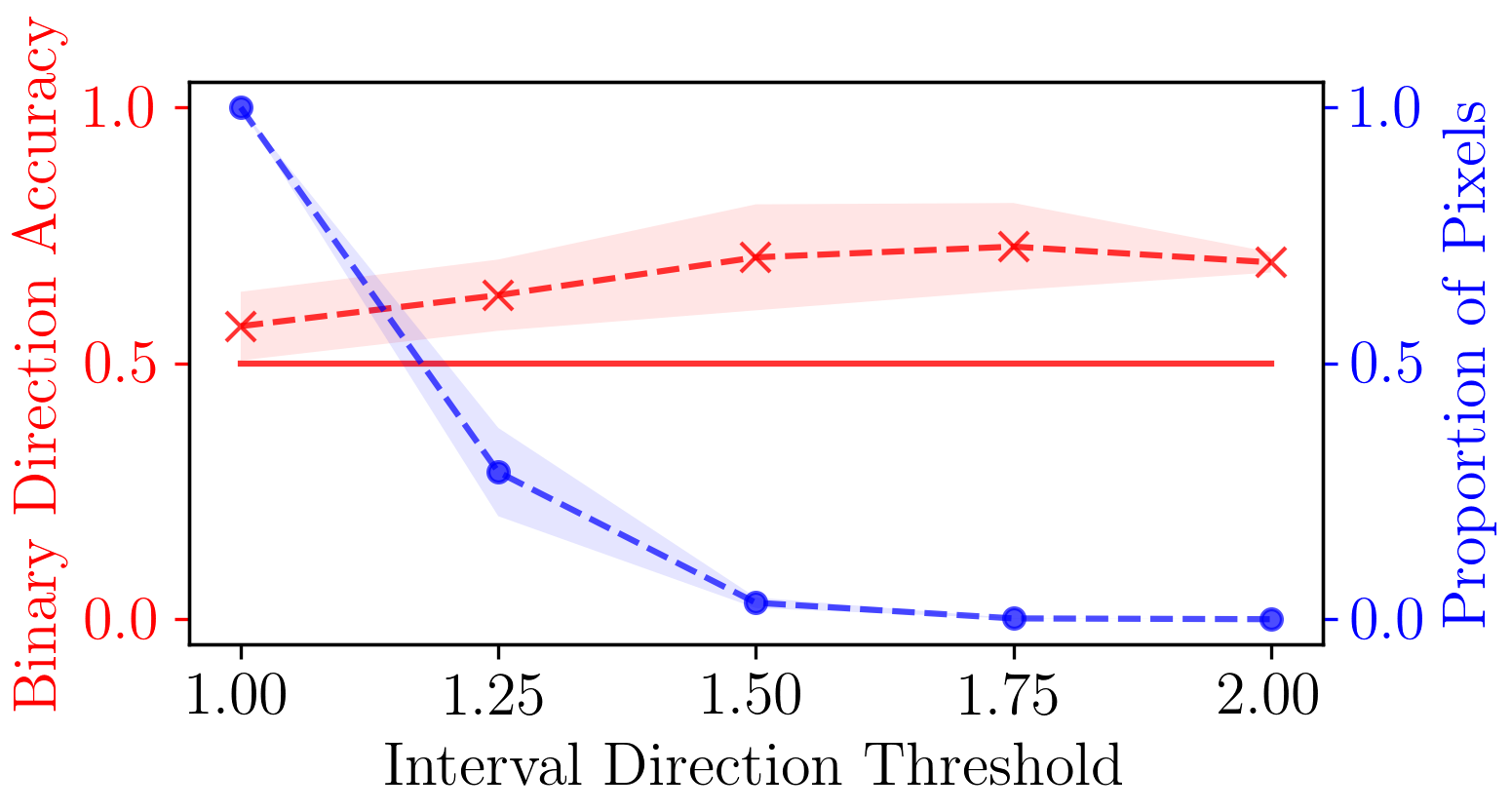}
    \caption{\textbf{Directional Information.} Interval direction thresholds are displayed on the $x$-axis. These are computed by dividing the larger interval half by the smaller half. Interval halves are computed relative to the point prediction. The left $y$-axis (in red) displays the direction accuracy which is the mean agreement between the interval directions and the actual position of the target relative to the prediction. Finally, the right $y$-axis (in blue) displays the proportion of pixels that are considered at a given threshold and accuracy evaluation. Results are shown for the \DeconvShort\ task (top graph) and the \CTShort\ task (bottom graph). \CTShort\ results are means and standard deviations across the four experimental runs.}
    \label{fig:direction}
\end{figure}

Third, as both interval bounds are optimized using the penalty term for interval size, the ground truth is generally distributed symmetrically inside the interval. Therefore, if the prediction from the underlying network lies closer to one boundary of the output interval, one can infer that the probability of the ground truth lying on the other side of the interval is higher. A quantitative assessment of this capability on the \DeconvShort\ task can be found in the top graph of \Cref{fig:direction}. The directional information contained in \IntervalShort\ uncertainty scores leads to direction accuracy that is 12 to 25 percentage points above chance. This is in contrast to symmetric uncertainty score approaches like \DropShort\ and \ProbShort. Beyond these defining features of the \IntervalShort\ method, practical considerations to take note of include run time and stability during training. Given an underlying DNN that requires $K$ operations for a single forward pass the following costs are incurred for obtaining uncertainty scores.
 \IntervalShort s require $2 \times K$ operations. \DropShort\ uses $T \times K$ operations where $T$ is the number of times the dropout distribution is sampled and \cite{gal_dropout_2016} use $T=10$ as a rule of thumb. \ProbShort\ needs $C + K$ operations, where $C$ is the number of operations required to compute the final output layer of the DNN. Assuming a modestly sized DNN with $K\gg C$ we can see that \IntervalShort s are situated between the very efficient \ProbShort\ and the more expensive \DropShort. With respect to stability during training we note that for very deep INNs, issues can arise due to the possibility of exponential growth of the intervals. This can be dealt with by choosing a small learning rate in such a situation or only training the interval parameters in the last several layers.
\vspace{-0.4cm}
\section{Uncertainties as Error Proxies}
\label{sec:Experiments}
We showcased the mechanics behind the \IntervalShort\ method and highlighted its unique capabilities regarding coverage bounds, noise behavior and directional information. In the following experiment, the \IntervalShort\ method for UQ is subjected to real-life data in direct comparison to two other, popular UQ methods, namely \DropShort\ and \ProbShort. As \DropShort\ and \ProbShort\ do not offer the above capabilities of \IntervalShort s the comparison is done with respect to an important use-case of UQ that allows for direct comparison: error correlation. Our \ErrorLong\ (\ErrorShort) experiment assesses how well uncertainty scores correspond to the prediction error of the model.
\subsection{Data}
\label{seq:data}
As we outlined in the introduction, transparency-enhancing tools can play an important role in making the prowess of DNNs more viable in sensitive application settings such as medical imaging workflows. Hence, the \ErrorLong\ experiment is performed on real-life CT data. 
Recently proposed deep learning reconstruction methods have shown promising results, effectively mitigating reconstruction artifacts caused by the ill-posedness of \Cref{eq:inv_prob}. Besides a few exceptions, most approaches assist the neural network by incorporating knowledge of the forward model $\bfA$, see \cite{arridge_maass_oektem_schoenlieb_2019} for an overview. In the \ErrorLong\ experiment, we focus on an ad-hoc technique, which first computes a classical solution via the filtered backprojection algorithm (FBP) \cite{natterer2001}. Depending on the degree of ill-posedness the FBP image is degraded by artifacts and missing edge information. The image is subsequently post-processed by a DNN in order to remove streaking artifacts and filling in the ``invisible'' edges \cite{jin_deep_2017,Bubba_2019}. We wish to emphasize that the result $\widetilde{\bfy}$ of such a post-processing of the FBP is generally not consistent with the measurements (i.e.\ $\bfx\neq \bfA\widetilde{\bfy}$). Hence, it does also not correspond to the desired image $\bfy$, making further information about the uncertainty of the network prediction much desirable. For our study of limited angle CT we consider a data set consisting of $512\times 512$ human CT scans from the AAPM Low Dose CT Grand Challenge\footnote{\url{https://www.aapm.org/GrandChallenge/LowDoseCT/}; All authors would like to acknowledge Dr. Cynthia McCollough, the
Mayo Clinic, and the American Association of Physicists in Medicine as well as the grants EB017095 and
EB017185 from the National Institute of Biomedical Imaging and Bioengineering for providing the AAPM data.} data \cite{mayo}.
\begin{figure}
    \setlength{\fboxsep}{0pt}%
    \setlength{\fboxrule}{1pt}%
    \begin{tabular}{cc}
         Input & Target\\
         \fbox{\includegraphics[width=0.28\textwidth]{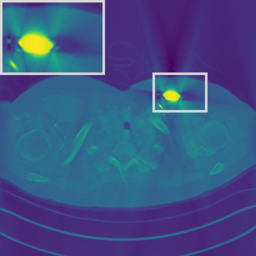}} &
         \fbox{\includegraphics[width=0.28\textwidth]{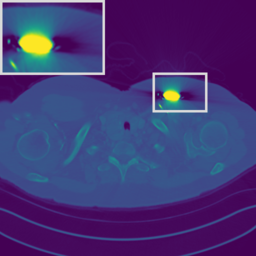}}
    \end{tabular}
    \vspace{0.15cm}
    \hspace*{0.05\textwidth}
    \includegraphics[width=0.9466\textwidth]{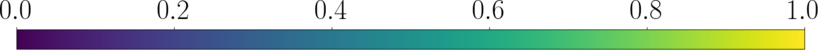}
    
    \begin{tabular}{rccc}
         &Dropout & Probout & Interval \\
         \rotatebox{90}{Prediction}&\fbox{\includegraphics[width=0.28\textwidth]{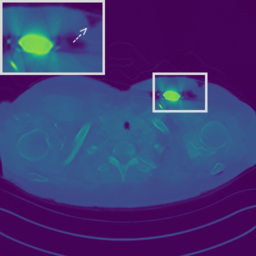}} &
         \fbox{\includegraphics[width=0.28\textwidth]{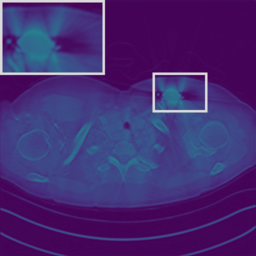}} &
         \fbox{\includegraphics[width=0.28\textwidth]{figures/CTImage/dropoutimg_output_arrow.png}} \\
         \rotatebox{90}{Uncertainty}&\fbox{\includegraphics[width=0.28\textwidth]{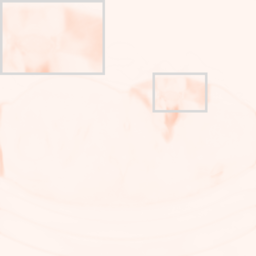}} &
         \fbox{\includegraphics[width=0.28\textwidth]{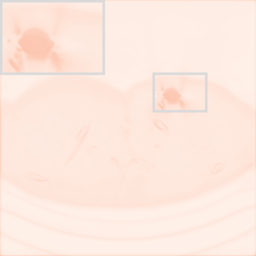}} &
         \fbox{\includegraphics[width=0.28\textwidth]{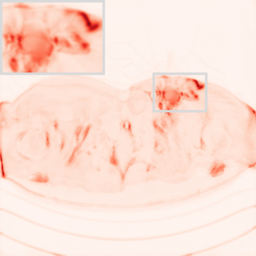}} \\
         \rotatebox{90}{Absolute Error}&\fbox{\includegraphics[width=0.28\textwidth]{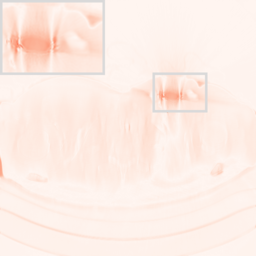}} &
         \fbox{\includegraphics[width=0.28\textwidth]{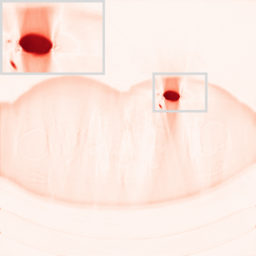}} &
         \fbox{\includegraphics[width=0.28\textwidth]{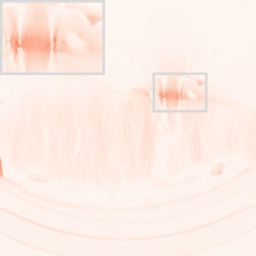}}
    \end{tabular}\\
    
    \vspace*{0.1cm}
    \hspace*{0.05\textwidth}
    \includegraphics[width=0.9466\textwidth]{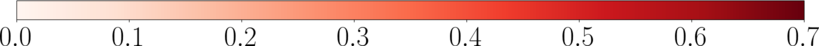}
    \caption{\textbf{\textsc{CT} Task Results.} Top row displays input (left) and corresponding target (right). Corresponding predictions (second row), uncertainty scores as standard deviation (\DropShort\ and \ProbShort) and interval size (\IntervalShort) (third row) as well as absolute errors (fourth row) are displayed below for each of the uncertainty methods.}
    \label{fig:ellipse_example_images1}
\end{figure}
It contains $2580$ images from $10$ patients in total. Eight of these ten patients were used for training ($2036$ samples), one for validation ($214$ samples) and one for testing ($330$ samples). The Radon transform and subsequent FBP were computed with a missing wedge of thirty degrees in the sinogram.
\subsection{Neural Network Architectures and Training Procedures}
For the \ErrorLong\ experiment the prediction DNN underlying all UQ methods is a U-Net as in \cite{ronneberger_u-net:_2015}. We added dropout with drop probability $0.7$ after each down-sampling and up-sampling layer. It was trained for $400$ epochs using Adam \cite{kingma2014adam} with a learning rate of $7.5\cdot 10^{-5}$ and mini-batches of size $12$. The \IntervalShort\ parameters were trained for $15$ epochs using Adam with learning rate $10^{-6}$ with $\beta=10^{-4}$ and mini-batches of size $6$. We limited the interval training to the last twelve layers. For \ProbShort\ we use the same U-Net architecture with an additional output channel for the uncertainty scores. It was initialized using the trained underlying U-Net and further trained for $400$ epochs using Adam with learning rate $10^{-7}$ and mini-batches of size $12$. We used $128$ forward passes to determine the mean and variance for \DropShort.
\subsection{Experiment}
We repeat the training four times for each UQ method and evaluate their effectiveness as an error proxy. The \textit{performance weighted correlation coefficient} (PWCC) of the uncertainty scores of each UQ method and the absolute prediction errors are compared. Performance weighted means the correlation coefficient is weighted by the mean squared error. This is necessary to discourage rewards for poor prediction models with high uncertainties everywhere. Not doing so would entail high correspondence between the uncertainty scores and prediction errors of prediction models that perform very poorly, which is not desirable. The final scores are computed as the mean correlation for all elements of the test set for all of the four training runs. The interval size (\IntervalShort) and standard deviation (\DropShort\ and \ProbShort) in the output are used as uncertainty scores. For a datapoint $(\bfx_i,\bfy_i)$ and a corresponding uncertainty map $\bfu_i$, the performance weighted correlation coefficient (PWCC) is thus computed as follows:
\begin{equation*}
\text{PWCC}(\bfx_i, \bfy_i, \bfu_i) = \frac{\text{corr}\left(|\bfPhi(\bfx_i)-\bfy_i|, \bfu_i\right)}{\text{MSE}(\bfPhi(\bfx_i),\bfy_i)}.
\end{equation*}
\begin{table}[t]
\caption{\textbf{EC Experiment Results}. Results of the Error Correlation experiment on the human CT data from the  AAPM Low Dose CT GrandChallenge. Each measure is reported as the mean over all data points and test runs. Numbers after the $\pm$ signify the standard deviation over the four experimental runs.}
\label{tab:results}
\begin{center}
\begin{small}
\begin{sc}

\begin{tabular}{lcc}
\toprule
Method & Test PWCC & Test MSE \\
\midrule
Dropout & $2170 \pm 513$ & $7.4 \pm 0.65 \times 10^{-4}$\\
ProbOut & $190 \pm 28$ & $6.7 \pm 2 \times 10^{-3}$\\
Interval & $\mathbf{2211 \pm 403}$ & $7.4 \pm 0.65 \times 10^{-4}$\\
\bottomrule
\end{tabular}

\end{sc}
\end{small}
\end{center}
\end{table}
\subsection{Results}
In \Cref{tab:results} the results for the error correlation experiment are documented. Each method for the delivery of uncertainty scores was evaluated on the errors of the prediction model on the test data. An example from this experiment can be seen in \Cref{fig:ellipse_example_images1}. First, we can observe the relatively poor prediction accuracy of the \ProbShort\ method compared to the underlying networks for the \DropShort\ and \IntervalShort\ methods, especially in the marked area in the top right. We can also see that both \IntervalShort\ and \DropShort\ are able to detect the error prone region in the top right, as well as an edge artifact produced by the prediction network (see arrow). Apart from this, the \IntervalShort\ method also highlights other regions in the image with high local intensity variations. In addition, we can again observe the \IntervalShort s performance with respect to coverage and directional information. A total $76 \pm 6 \%$ of test targets in the \CTShort\ data are contained in the produced intervals. In the bottom graph of \Cref{fig:direction} the directional accuracy of the \IntervalShort\ becomes more pronounced from $57\%$ up to $72\%$ as the interval direction threshold grows.
\section{Conclusion}
We introduced a novel, non-Bayesian method based on interval propagation for computing upper and lower bounds and subsequent uncertainty maps for pre-trained neural networks. We presented the advantages of this method and compared it to two other existing methods. We analyzed these methods on a CT reconstruction data set with regard to their correspondence with the absolute error and the sufficiency of their explanations. We found that interval neural networks not only gave the most consistent results, but also grant additional insights concerning coverage of the intervals, noise behaviour and directional information for the ground truth.
\clearpage
\bibliography{references/luis_refs,references/related_work,references/janmax_refs}
\bibliographystyle{amsplain}
\clearpage
\begin{appendices}
\crefalias{section}{appsec} 
\section*{Appendices}
\begin{figure*}[t]
    \section{\DeconvShort\ Experiments: Additional Samples}
    \label{sec:1dextra}
    \centering
    \begin{tabular}{rrr}
        &\includegraphics[width=0.425\textwidth]{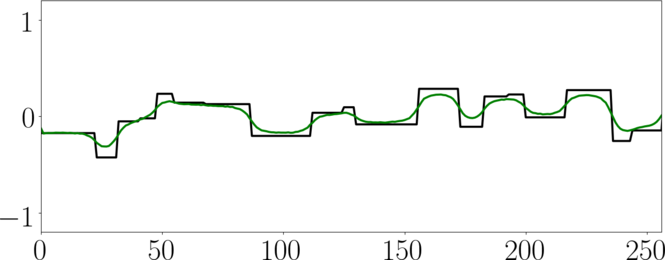}&
        \includegraphics[width=0.425\textwidth]{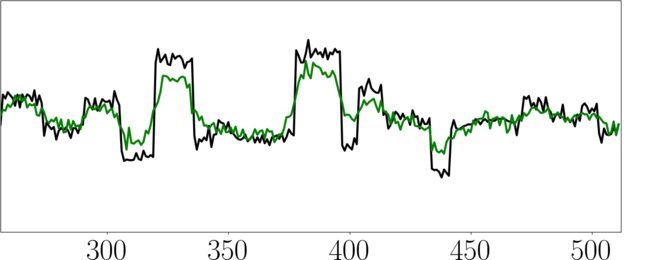} \\
        
        \multirow[t]{2}{*}{(I)$\begin{cases}$\vspace*{3.3cm}$\end{cases}$ \hspace*{-0.8cm}}
        &\includegraphics[width=0.425\textwidth]{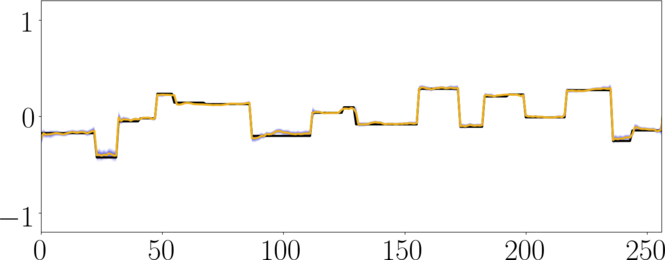}& 
        \includegraphics[width=0.425\textwidth]{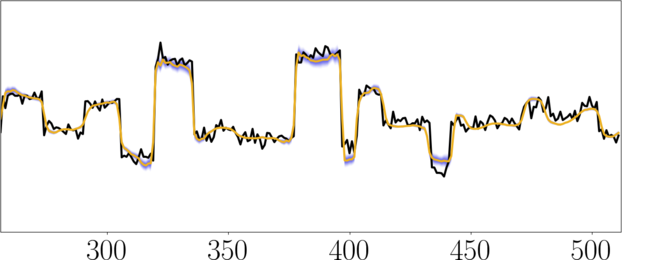} \\
        
        &\includegraphics[width=0.425\textwidth]{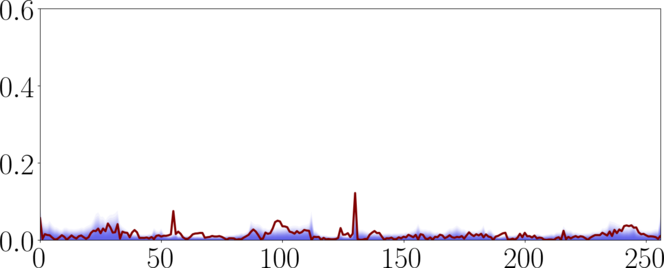}& 
        \includegraphics[width=0.425\textwidth]{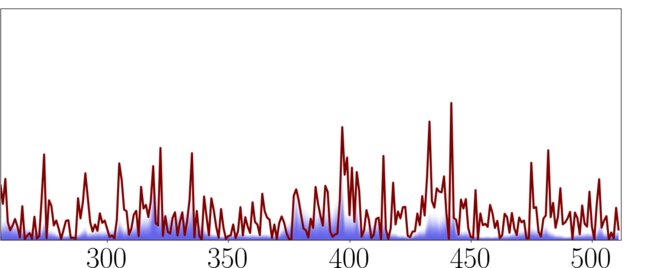} \\
        
        \multirow[t]{2}{*}{(II)$\begin{cases}$\vspace*{3.3cm}$\end{cases}$ \hspace*{-0.8cm}}
        &\includegraphics[width=0.425\textwidth]{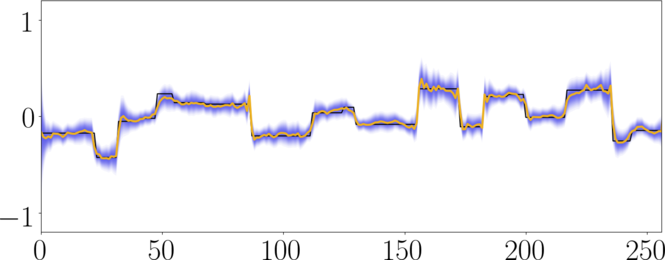}& 
        \includegraphics[width=0.425\textwidth]{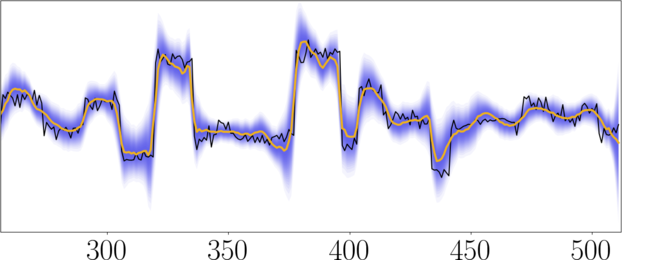} \\
        
        &\includegraphics[width=0.425\textwidth]{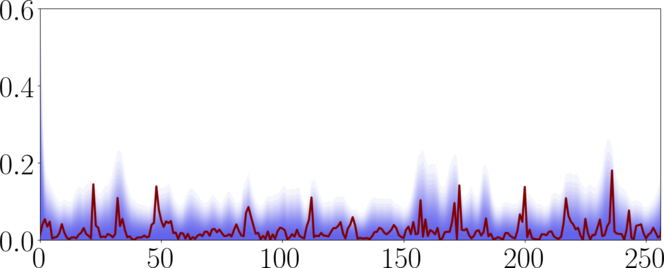}&
        \includegraphics[width=0.425\textwidth]{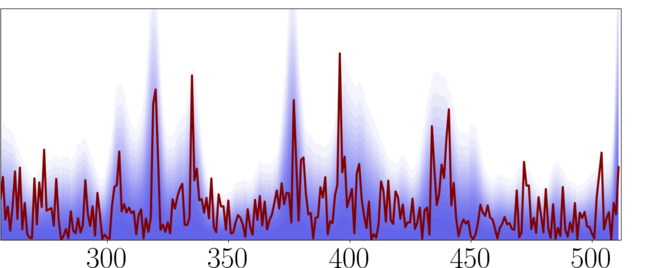} \\
        
        \multirow[t]{2}{*}{(III)$\begin{cases}$\vspace*{3.3cm}$\end{cases}$ \hspace*{-0.8cm}}
        &\includegraphics[width=0.425\textwidth]{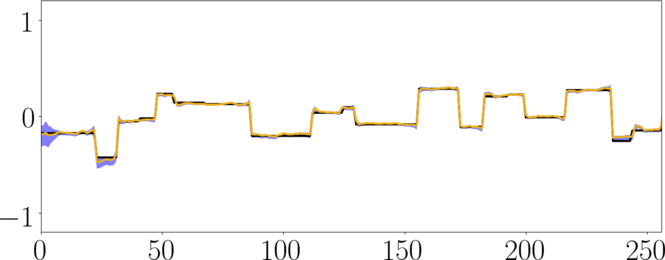}&
        \includegraphics[width=0.425\textwidth]{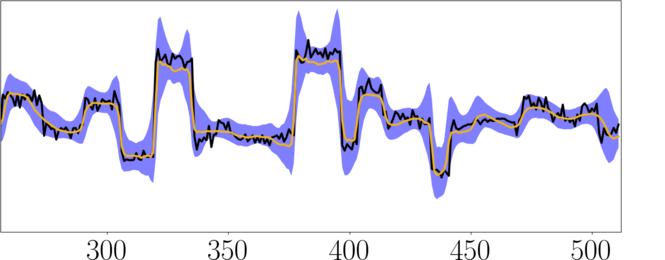} \\
        &\includegraphics[width=0.425\textwidth]{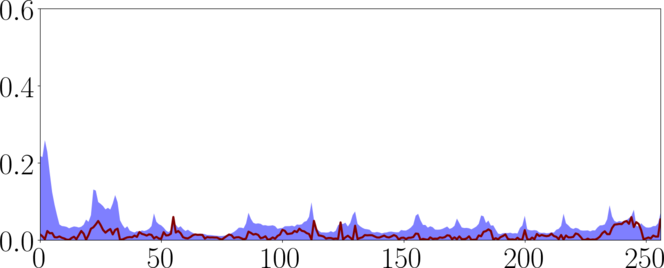}& 
        \includegraphics[width=0.425\textwidth]{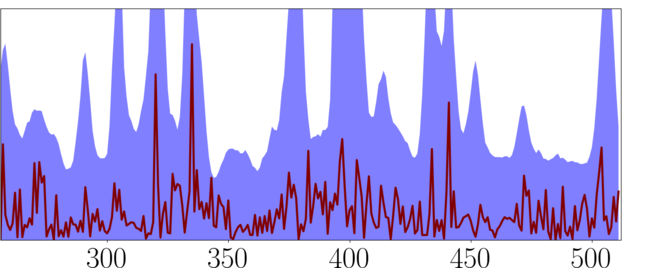} \\
        
        &\multicolumn{2}{l}{\hspace{0.026\textwidth}\includegraphics[width=0.77\textwidth]{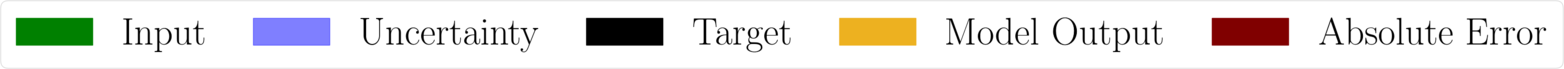}}
    \end{tabular}
    \caption{\textbf{\textsc{1DDeconv} Task Results.} Results for the \textsc{1DDeconv} task on the same sample without noise on the left and with Gaussian noise ($\sigma=0.05$) on the right. The first row displays input and target vector. The figures below show target, network output, together with the uncertainty estimation for the upper graph and the uncertainty estimation plotted against the absolute error in the lower graph for each corresponding method; (I) \textsc{MCDrop}, (II) \textsc{ProbOut}, (III) \textsc{INN}.}
\end{figure*}

\begin{figure}
    \section{\CTShort\ Experiments: Additional Samples}
    \label{sec:ctextra}
    \setlength{\fboxsep}{0pt}%
    \setlength{\fboxrule}{1pt}%
    \begin{tabular}{cc}
         Input & Target\\
         \fbox{\includegraphics[width=0.28\textwidth]{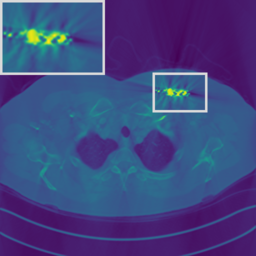}} &
         \fbox{\includegraphics[width=0.28\textwidth]{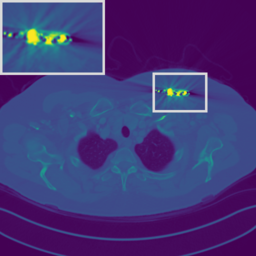}}
    \end{tabular}
    \centering
    \vspace{0.15cm}
    \hspace*{0.05\textwidth}
    \includegraphics[width=0.9466\textwidth]{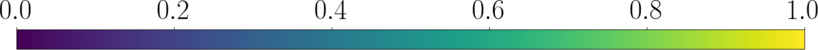}
    
    \begin{tabular}{rccc}
         &Dropout & Probout & Interval \\
         \rotatebox{90}{Prediction}&\fbox{\includegraphics[width=0.28\textwidth]{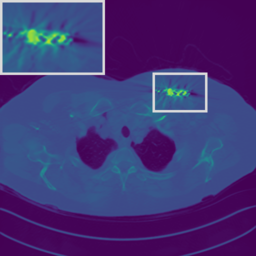}} &
         \fbox{\includegraphics[width=0.28\textwidth]{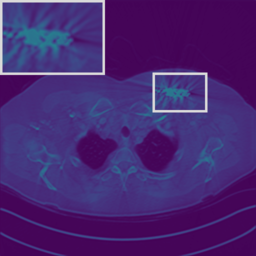}} &
         \fbox{\includegraphics[width=0.28\textwidth]{figures/CTImage1/dropoutimg_output.png}} \\
         \rotatebox{90}{Uncertainty}&\fbox{\includegraphics[width=0.28\textwidth]{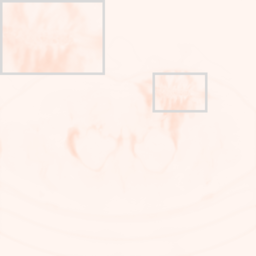}} &
         \fbox{\includegraphics[width=0.28\textwidth]{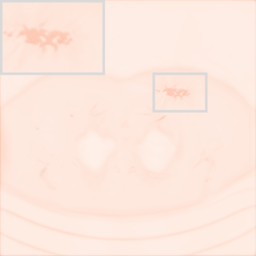}} &
         \fbox{\includegraphics[width=0.28\textwidth]{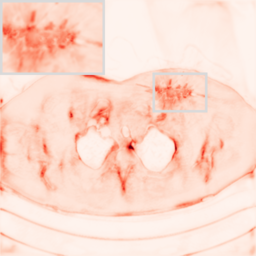}} \\
         \rotatebox{90}{Absolute Error}&\fbox{\includegraphics[width=0.28\textwidth]{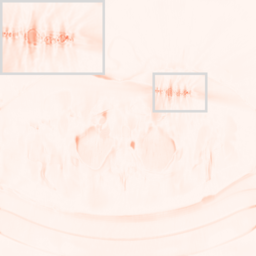}} &
         \fbox{\includegraphics[width=0.28\textwidth]{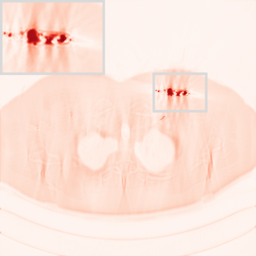}} &
         \fbox{\includegraphics[width=0.28\textwidth]{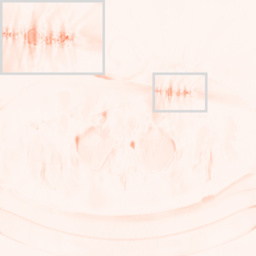}}
    \end{tabular}\\
    
    \vspace*{0.1cm}
    \hspace*{0.05\textwidth}
    \includegraphics[width=0.9466\textwidth]{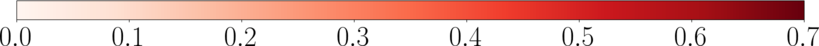}
    
    \caption{\textbf{\textsc{CT} Task Results.} Top row displays input (left) and corresponding target (right). Corresponding predictions (second row), uncertainty scores as standard deviation (\DropShort\ and \ProbShort) and interval size (\IntervalShort) (third row) as well as absolute errors (fourth row) are displayed below for each of the uncertainty methods.}
\end{figure}

This supplement to the paper \textit{Interval Neural Networks: Uncertainty Scores} contains the following four addenda: a derivation of the formal coverage bound argument from the main paper (\Cref{sec:markov}), a Bayesian treatment of \IntervalShort s for an additional perspective (\Cref{sec:bayes}) and, finally, three more randomly sampled results from the experiments on the \DeconvLong\ (\Cref{sec:1dextra}) and \CTLong\ (\Cref{sec:ctextra}) data sets to facilitate a broader qualitative assessment. 
\section{\IntervalShort\ Coverage Bound}
\label{sec:markov}
For both data sets in the main paper the proportion of ground truth values that lie inside the intervals were documented. Furthermore, it was argued that this type of coverage bound can be theoretically justified using the Markov Inequality. In the following this argument from the main paper is formally derived. 
\par 
For some data distribution $X,Y$ and a tightness parameter $\beta$ the following loss is used:

\begin{align*}
    \CL(\underline{\Phi},\overline{\Phi}) &= \E\left[ \max(\bfy-\overline{\Phi}(\bfx), 0)^2 + \max(\underline{\Phi}(\bfx)-\bfy, 0)^2 + \beta \cdot (\overline{\Phi}(\bfx)-\underline{\Phi}(\bfx)) \right] \\
    &= \int_\CX \E\left[ \max(\bfy-\overline{\Phi}(\bfx), 0)^2 \,\middle|\, \bfx\right] + \E\left[ \max(\underline{\Phi}(\bfx)-\bfy, 0)^2 \,\middle|\, \bfx\right] \\
    &\qquad +   \beta \cdot (\overline{\Phi}(\bfx)-\underline{\Phi}(\bfx)) \,\D\P_X(\bfx).
\end{align*}
Assuming that this loss is optimized during training yields
\begin{align*}
0 &= \int_\CX \frac{\partial }{\partial \overline{\Phi}(\bfx)} \big( \E\left[ \max(\bfy-\overline{\Phi}(\bfx), 0)^2 \,\middle|\, \bfx\right] + \E\left[ \max(\underline{\Phi}(\bfx)-\bfy, 0)^2 \,\middle|\, \bfx\right] \\
&\qquad+   \beta \cdot (\overline{\Phi}(\bfx)-\underline{\Phi}(\bfx)) \big) \ \D\P_X(\bfx) \\
\Longleftrightarrow \quad \ \ \ 0 &= -\int_\CX 2\E\big[\max(\bfy-\overline{\Phi}(\bfx), 0)\big]\ \D\P_X(\bfx)+\beta\\
\Longleftrightarrow \quad \frac{1}{2}\beta &= \int_\CX \E\big[\max(\bfy-\overline{\Phi}(\bfx), 0)\big]\ \D\P_X(\bfx)
\intertext{and analogously}
\frac{1}{2}\beta &= \int_\CX \E\big[\max(\underline{\Phi}(\bfx)-\bfy, 0)\big]\ \D\P_X(\bfx).
\end{align*}

Using the Markov Inequality with $h_1(\zeta):= \text{max}(\zeta-\overline{\Phi}(\bfx),0)$ and $h_2(\zeta):= \text{max}(\zeta+\underline{\Phi}(\bfx),0)$, we obtain that for the marginalized distribution the following holds true:
\begin{equation*}
\P(\bfy \geq \overline{\Phi}(\bfx)+\lambda \beta)  \leq \frac{\E\big[h_1(\bfy) \big]}{h_1(\overline{\Phi}(\bfx)+\lambda \beta)} = \frac{\E\big[\max(\bfy-\overline{\Phi}(\bfx), 0) \big]}{\lambda \beta}
\end{equation*}
and
\begin{align*}
 \P(\bfy \leq \underline{\Phi}(\bfx)-\lambda \beta) & = \P(-\bfy \geq -\underline{\Phi}(\bfx)+\lambda \beta) & \\ 
 &\qquad \qquad \leq \frac{\E\big[h_2(-\bfy) \big]}{h_2(-\underline{\Phi}(\bfx)+\lambda \beta)} = \frac{\E\big[\text{max}(\underline{\Phi}(\bfx)-\bfy,0) \big]}{\lambda \beta}.&
\end{align*}
Hence, with $A = \big\lbrace\text{Label is inside interval bounds plus $\lambda \beta$} \big\rbrace$ we conclude that
\begin{align*}
\P(A) &=
\int_X \P(\underline{\Phi}(\bfx)-\lambda \beta\leq \bfy \leq \overline{\Phi}(\bfx)+\lambda \beta)\ d\P_X \\
&= 1-\int_X \P(\bfy \leq \underline{\Phi}(\bfx)-\lambda \beta)+ \P(\bfy \geq \overline{\Phi}(\bfx)+\lambda \beta)\ d\P_X \\
& \geq 1- \int_X \frac{\E\big[\text{max}(\bfy-\overline{\Phi}(\bfx),0) \big]}{\lambda \beta}+\frac{\E\big[\text{max}(\underline{\Phi}(\bfx)-\bfy,0) \big]}{\lambda \beta} d\P_X\\
& = 1- \frac{1}{\lambda}.
\end{align*}

We can furthermore bound the probability that for a given data point $\bfx$ the label $\bfy$ has a probability of more than $\alpha$ to be outside the interval bounds:
\begin{alignat*}{2}
&&\frac{1}{\lambda} &\geq \E_X\big[\P( \bfy < \underline{\Phi}(\bfx)-\lambda \beta,\ \bfy > \overline{\Phi}(\bfx)+\lambda \beta) \big] \\
&&& \qquad \qquad \geq \E_X\big[\alpha \1_{\P( \bfy < \underline{\Phi}(\bfx)-\lambda \beta,\ \bfy > \overline{\Phi}(\bfx)+\lambda \beta)> \alpha} \big]\\
&\Longleftrightarrow \qquad &\frac{1}{\lambda \alpha} &\geq \E_X\big[\1_{\P( \bfy < \underline{\Phi}(\bfx)-\lambda \beta,\ \bfy > \overline{\Phi}(\bfx)+\lambda \beta)> \alpha} \big] \\
&&&\qquad \qquad= 1-\E_X\big[\1_{\P(\underline{\Phi}(\bfx)-\lambda \beta\leq \bfy \leq \overline{\Phi}(\bfx)+\lambda \beta)\geq 1-\alpha} \big]\\
&\Longleftrightarrow \qquad &1-\frac{1}{\lambda \alpha} &\leq \E_X\big[\1_{\P(\underline{\Phi}(\bfx)-\lambda \beta\leq \bfy \leq \overline{\Phi}(\bfx)+\lambda \beta)\geq 1-\alpha} \big].
\end{alignat*}
In words, for $\lambda>0$ and $\alpha>0$ the probability mass of all samples $\bfx$, for which the corresponding label $\bfy$ has the probability of at least $1-\alpha$ to be inside the interval, is at least $ 1-\frac{1}{\lambda \alpha}$.
\section{\IntervalShort s and the Bayesian View}
\label{sec:bayes}
As described in Section 2 on related work, popular UQ approaches for neural networks have their roots in a Bayesian treatment of the learning problem. In a nutshell, this involves modelling the unknown data distribution $(X,Y)$ on $\CX\times\CY$ via a neural network $\Phi_W\colon\CX\to\CY$, where $W$ is now also a random variable and represents the collection of all network parameters. More precisely, one assumes that $p(Y\vert X, W)$ follows a simple distribution depending on $X$ and $W$ through $\Phi_W(X)$. A typical choice is a Gaussian distribution $Y\vert X,W \sim \CN(\Phi_W(X),\tau^{-2}I)$ with mean $\Phi_W(X)$ and some fixed precision $\tau$. The network training requires the estimation of 
\begin{equation}\label{eq:weight_post}
p(W\vert\bfX,\bfY) = \frac{p(\bfY\vert\bfX,W)p(W)}{\int p(\bfY\vert\bfX,W)p(W)\,\D W},
\end{equation}
from given training data $(\bfX,\bfY)=\{(\bfx_i,\bfy_i)\}_{i=1}^m$ and some prior assumption $p(W)$ on the network parameters. Inference for a new input $\bfx$ requires the estimation of the posterior
\begin{equation}\label{eq:post}
p(\bfy\vert\bfx,\bfX,\bfY) = \int p(\bfy\vert \bfx, W) p(W\vert \bfX, \bfY)\,\D W.
\end{equation}
The evidence, that is the denominator in \eqref{eq:weight_post}, is typically intractable. Variational Bayesian methods try to approximate $p(W\vert\bfX,\bfY)$ by another distribution $q_\theta(W)$ from a family of distributions $q_\theta$ parametrized by $\theta$. Minimizing the KL-divergence between $p(W\vert\bfX,\bfY)$ and $q_\theta(W)$ is equivalent to maximizing the evidence lower bound (ELBO)
\begin{equation}\label{eq:elbo}
    \int q_\theta(W)\log p(\bfY\vert \bfX,W)\,\D W - \KL(q_\theta(W)\vert\vert p(W)).
\end{equation}
Training the network in this variational setting entails finding the optimal parameter choice $\theta^\ast$ maximizing \eqref{eq:elbo}, and inference can be approximated by
\begin{equation}\label{eq:approx_post}
p(\bfy\vert\bfx,\bfX,\bfY) \approx \int p(\bfy\vert \bfx, W) q_{\theta^\ast}(W)\,\D W.
\end{equation}

In light of this tradition, we want to briefly demonstrate how our Interval Neural Networks can also be viewed within the framework of variational Bayesian networks.

For an $L$-layer neural network with weight matrices $\bfW^{(l)}$ and bias vectors $\bfb^{(l)}$, our interval network approach introduces upper and lower bound parameters $\theta=\{\underline{\bfW}^{(l)}, \overline{\bfW}^{(l)}, \underline{\bfb}^{(l)}, \overline{\bfb}^{(l)}\}_{l=1}^{L}$. But instead of precisely parametrizing the approximating distribution by $\theta$, we allow $q_\theta$ to be any distribution of weights and biases supported within the specified intervals. We now want to analyze the ELBO in \eqref{eq:elbo} and the approximate posterior in \eqref{eq:approx_post} in this situation.

Recall that, given the interval bounds $\theta$, the range of possible values of $\Phi_W(\bfx)$ for a fixed input $\bfx$ and $W$ distributed according to $q_\theta(W)$ is denoted as $[\underline{\Phi}(\bfx), \overline{\Phi}(\bfx)]$. Further, for any target $\bfy$ we denote the choice of weights achieving the best and worst approximation within this range as
\begin{equation*}
    \underline{\bfW}(\bfx,\bfy) = \argmin_{W\sim q_\theta} \|\Phi_W(\bfx)-\bfy\|_2^2
    \qquad\text{and}\qquad
    \overline{\bfW}(\bfx,\bfy) = \argmax_{W\sim q_\theta} \|\Phi_W(\bfx)-\bfy\|_2^2.
\end{equation*}
This allows us to estimate the first term in the ELBO as
\begin{equation*}
     \int q_\theta(W)\log p(\bfY\vert \bfX,W)\,\D W \leq -m\log(C) -  \sum_{i=1}^m \frac{\tau^{2d}}{2}\|\Phi_{\underline{\bfW}(\bfx_i,\bfy_i)}(\bfx_i)-\bfy_i\|_2^2,
\end{equation*}
where $C=(2\pi \tau^{-2})^{d/2}$ is the normalizing constant of the Gaussian density with precision $\tau$. Similarly
\begin{align*}
     \int q_\theta(W)\log p(\bfY\vert \bfX,W)\,\D W & \geq -m\log(C) -  \sum_{i=1}^m \frac{\tau^{2d}}{2}\|\Phi_{\overline{\bfW}(\bfx_i,\bfy_i)}(\bfx_i)-\bfy_i\|_2^2\\
     &\geq -m\log(C) -  \sum_{i=1}^m \tau^{2d} \big(\|\Phi_{\underline{\bfW}(\bfx_i,\bfy_i)}(\bfx_i)-\bfy_i\|_2^2 \\ &\qquad+\|\Phi_{\overline{\bfW}(\bfx_i,\bfy_i)}(\bfx_i)-\Phi_{\underline{\bfW}(\bfx_i,\bfy_i)}(\bfx_i) \|_2^2\big) \\
     & \geq -m\log(C) -  \sum_{i=1}^m \tau^{2d}\big(\|\Phi_{\underline{\bfW}(\bfx_i,\bfy_i)}(\bfx_i)-\bfy_i\|_2^2\\ 
     &\qquad+\|\overline{\Phi}(\bfx_i)-\underline{\Phi}(\bfx_i) \|_2^2\big).
\end{align*}
We observe that minimizing the INN loss $\CL(\underline{\Phi},\overline{\Phi})$ with $\beta=1$ corresponds to maximizing a lower bound for one part of the ELBO. The other part of the ELBO, the KL-divergence to the prior, corresponds to weight regularisation during the network training, e.g.\ weight decay. Further, the gap between the upper and lower bound on the ELBO is determined by $\sum_i \tau^{2d} \|\overline{\Phi}(\bfx_i)-\underline{\Phi}(\bfx_i) \|_2^2$. Therefore, the size of the output intervals also corresponds to how far from the true ELBO we are, when considering the training loss $\CL$ instead.

\begin{figure}
    \centering
    \input{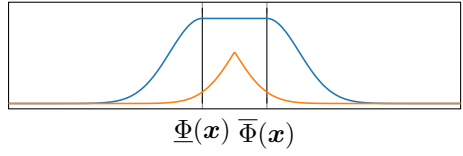}
    \caption{\textbf{\IntervalShort s and Bayes.} Schematic visualization of the lower and upper bounds for the predictive posterior  of INN inference derived from variational Bayesian principles. The INN prediction interval is marked by vertical lines.}
    \label{fig:post}
\end{figure}

During inference, the approximate posterior in \eqref{eq:approx_post} can then be estimated from the bounds
\begin{equation*}
\frac{1}{C} e^{-\frac{\tau^{2d}}{2} \|\Phi_{\overline{\bfW}(\bfx,\bfy)}(\bfx) - \bfy\|_2^2} \leq \int p(\bfy\vert \bfx, W) q_{\theta^\ast}(W)\,\D W \leq \frac{1}{C} e^{-\frac{\tau^{2d}}{2} \|\Phi_{\underline{\bfW}(\bfx,\bfy)}(\bfx) - \bfy\|_2^2}.
\end{equation*}
A schematic visualization of these bounds can be seen in \cref{fig:post}. Even though the true posterior can lie anywhere between the bounds, we observe a fast decay of the probability of the target $\bfy$ lying far outside the predicted interval $[\underline{\Phi}(\bfx), \overline{\Phi}(\bfx)]$. This is line with the findings derived via the Markov bound in \cref{sec:markov}.

\begin{figure*}[t]
    \centering
    \begin{tabular}{rrr}
        &\includegraphics[width=0.425\textwidth]{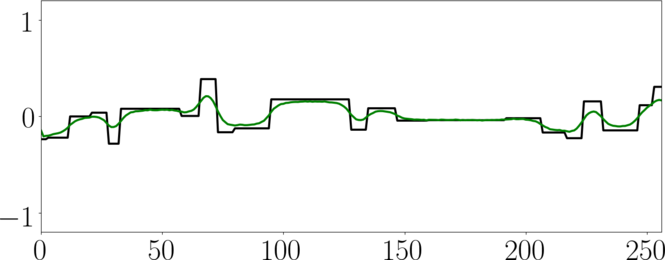}&
        \includegraphics[width=0.425\textwidth]{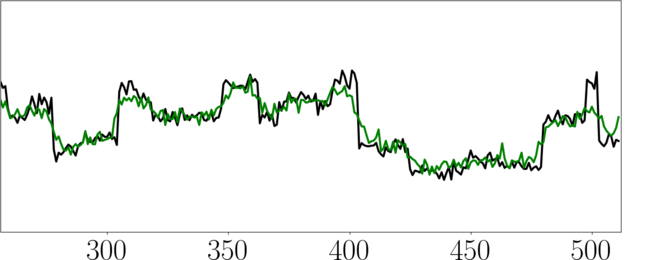} \\
        
        \multirow[t]{2}{*}{(I)$\begin{cases}$\vspace*{3.3cm}$\end{cases}$ \hspace*{-0.8cm}}
        &\includegraphics[width=0.425\textwidth]{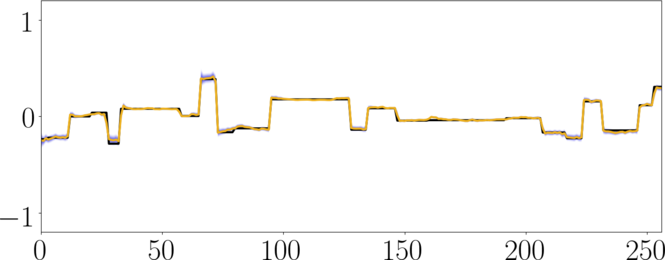}& 
        \includegraphics[width=0.425\textwidth]{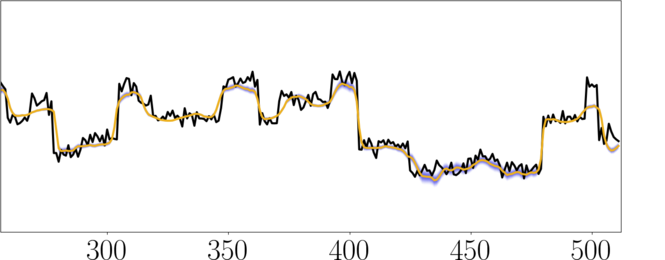} \\
        
        &\includegraphics[width=0.425\textwidth]{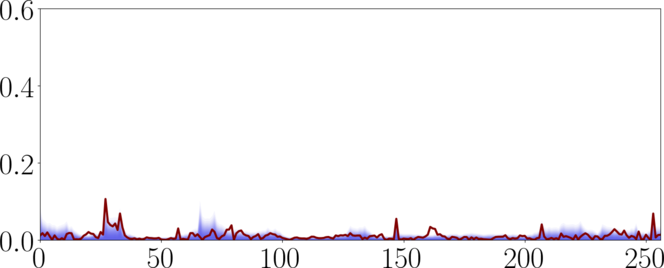}& 
        \includegraphics[width=0.425\textwidth]{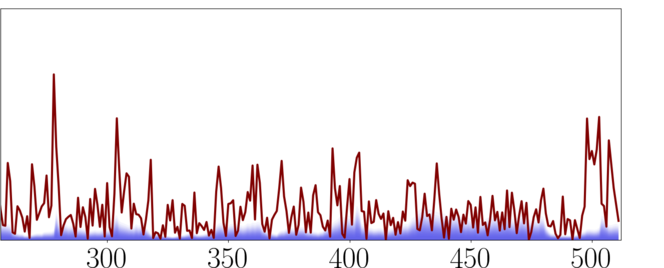} \\
        
        \multirow[t]{2}{*}{(II)$\begin{cases}$\vspace*{3.3cm}$\end{cases}$ \hspace*{-0.8cm}}
        &\includegraphics[width=0.425\textwidth]{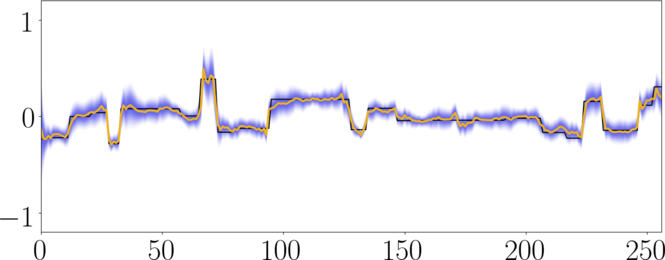}& 
        \includegraphics[width=0.425\textwidth]{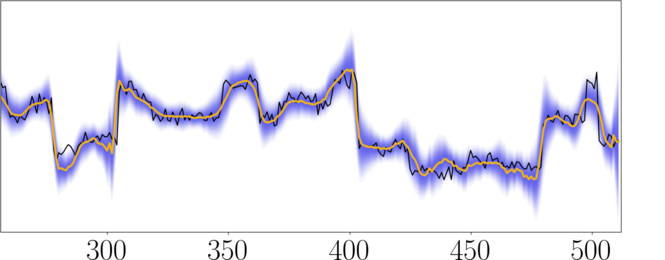} \\
        
        &\includegraphics[width=0.425\textwidth]{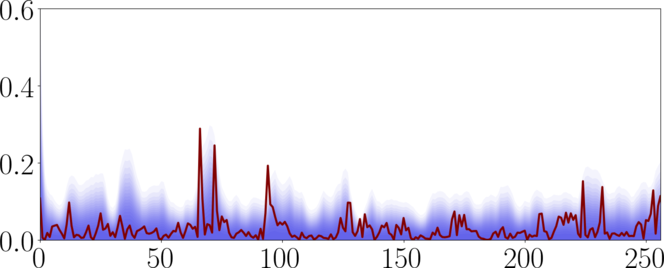}&
        \includegraphics[width=0.425\textwidth]{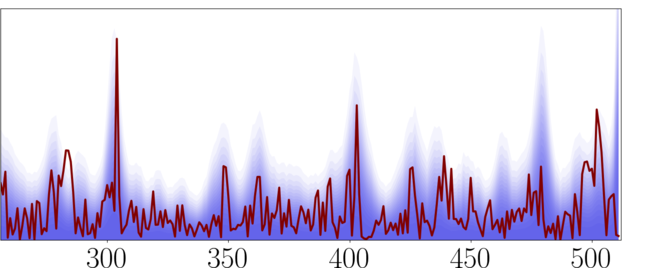} \\
        
        \multirow[t]{2}{*}{(III)$\begin{cases}$\vspace*{3.3cm}$\end{cases}$ \hspace*{-0.8cm}}
        &\includegraphics[width=0.425\textwidth]{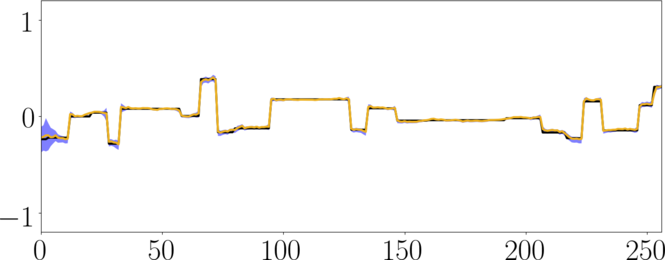}&
        \includegraphics[width=0.425\textwidth]{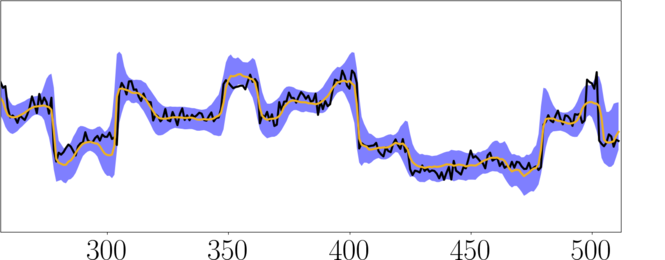} \\
        &\includegraphics[width=0.425\textwidth]{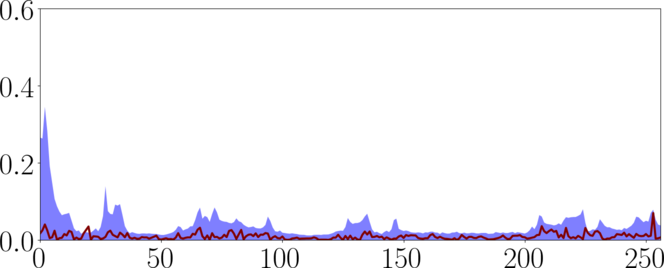}& 
        \includegraphics[width=0.425\textwidth]{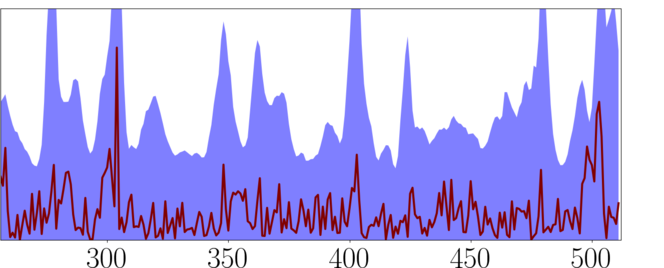} \\
        
        &\multicolumn{2}{l}{\hspace{0.026\textwidth}\includegraphics[width=0.77\textwidth]{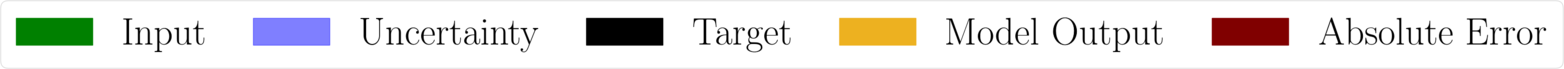}}
    \end{tabular}
    \caption{\textbf{\textsc{1DDeconv} Task Results.} Results for the \textsc{1DDeconv} task on the same sample without noise on the left and with Gaussian noise ($\sigma=0.05$) on the right. The first row displays input and target vector. The figures below show target, network output, together with the uncertainty estimation for the upper graph and the uncertainty estimation plotted against the absolute error in the lower graph for each corresponding method; (I) \textsc{MCDrop}, (II) \textsc{ProbOut}, (III) \textsc{INN}.}
\end{figure*}
\begin{figure*}[t]
    \centering
    \begin{tabular}{rrr}
        &\includegraphics[width=0.425\textwidth]{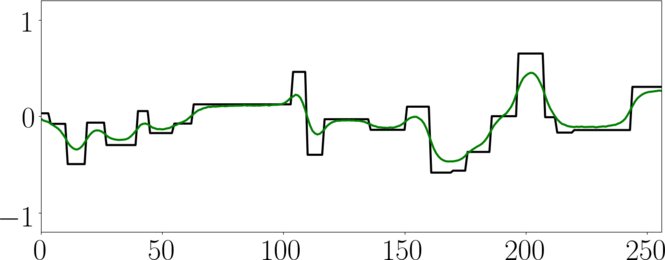}&
        \includegraphics[width=0.425\textwidth]{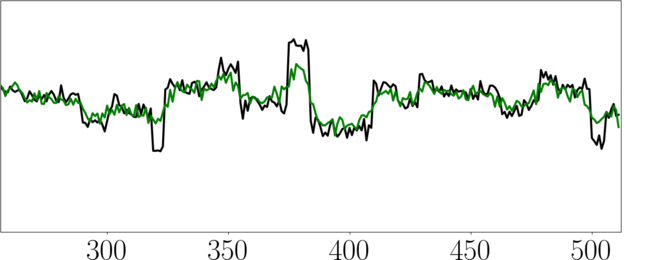} \\
        
        \multirow[t]{2}{*}{(I)$\begin{cases}$\vspace*{3.3cm}$\end{cases}$ \hspace*{-0.8cm}}
        &\includegraphics[width=0.425\textwidth]{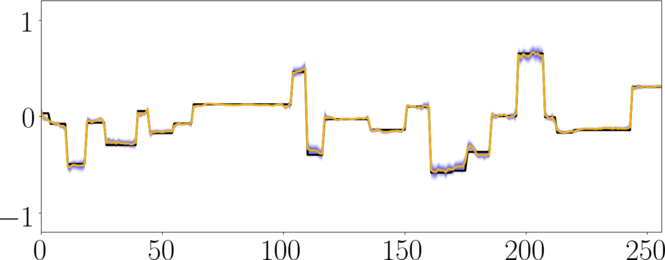}& 
        \includegraphics[width=0.425\textwidth]{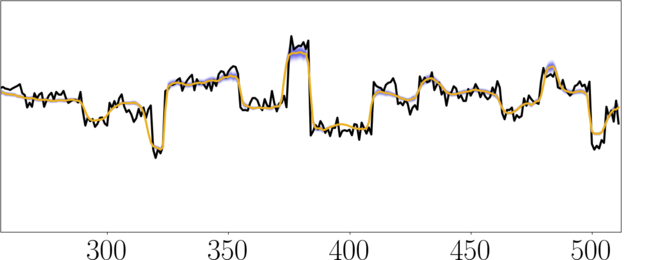} \\
        
        &\includegraphics[width=0.425\textwidth]{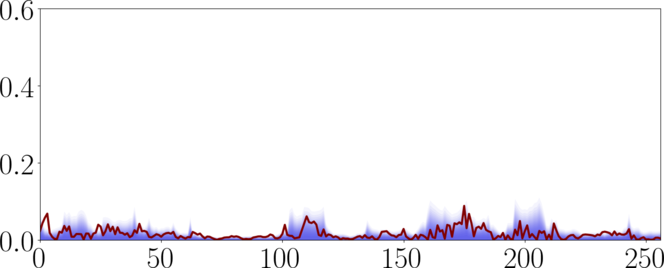}& 
        \includegraphics[width=0.425\textwidth]{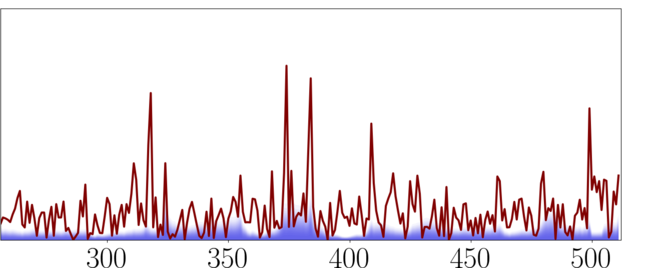} \\
        
        \multirow[t]{2}{*}{(II)$\begin{cases}$\vspace*{3.3cm}$\end{cases}$ \hspace*{-0.8cm}}
        &\includegraphics[width=0.425\textwidth]{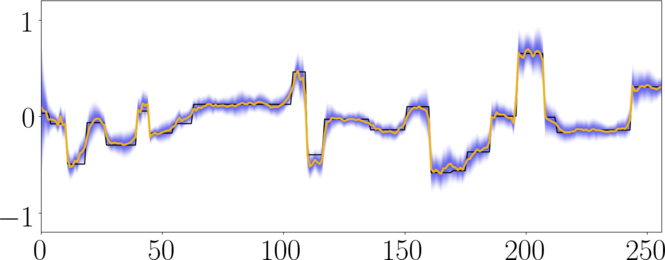}& 
        \includegraphics[width=0.425\textwidth]{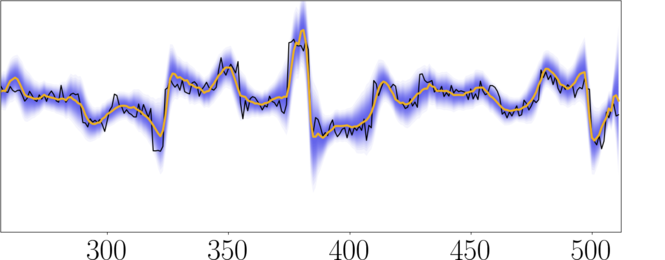} \\
        
        &\includegraphics[width=0.425\textwidth]{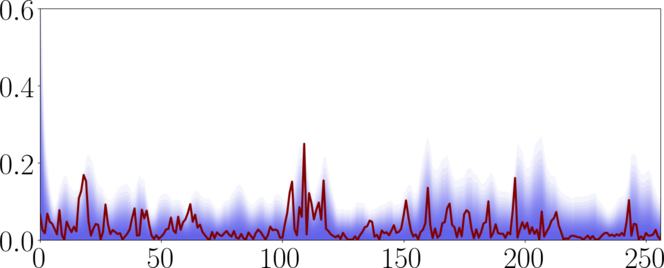}&
        \includegraphics[width=0.425\textwidth]{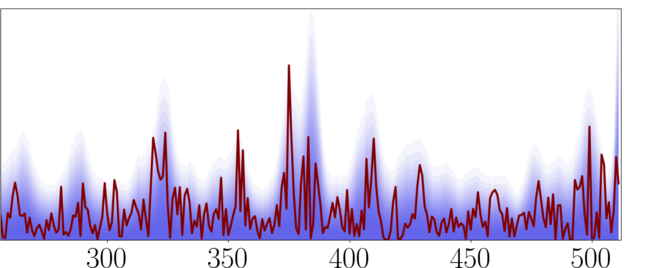} \\
        
        \multirow[t]{2}{*}{(III)$\begin{cases}$\vspace*{3.3cm}$\end{cases}$ \hspace*{-0.8cm}}
        &\includegraphics[width=0.425\textwidth]{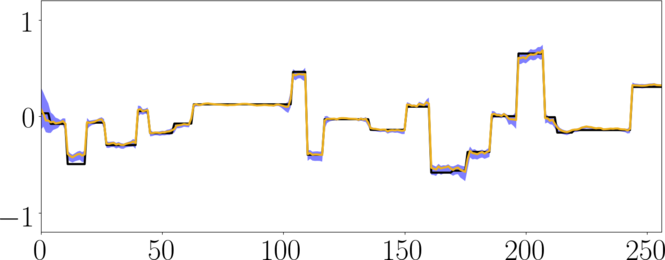}&
        \includegraphics[width=0.425\textwidth]{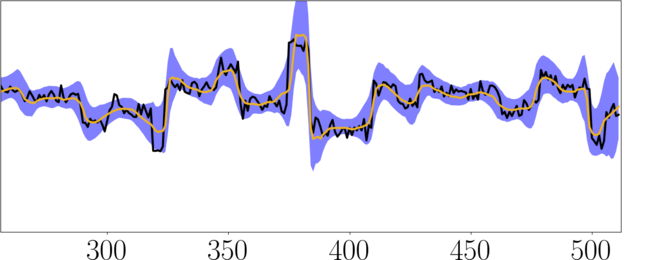} \\
        &\includegraphics[width=0.425\textwidth]{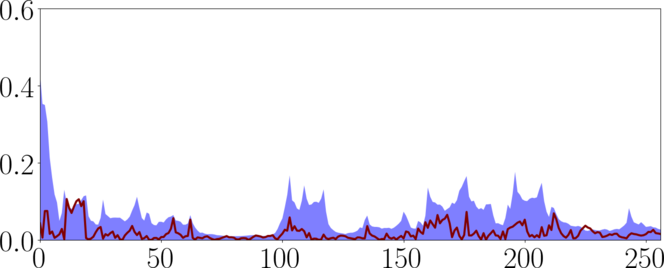}& 
        \includegraphics[width=0.425\textwidth]{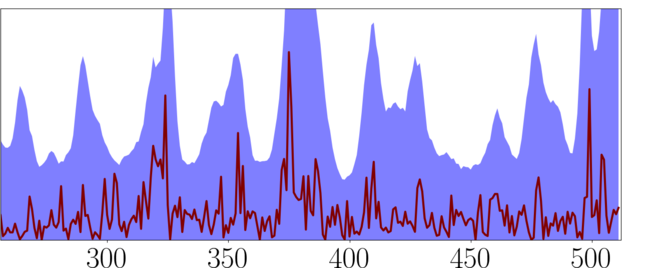} \\
        
        &\multicolumn{2}{l}{\hspace{0.026\textwidth}\includegraphics[width=0.77\textwidth]{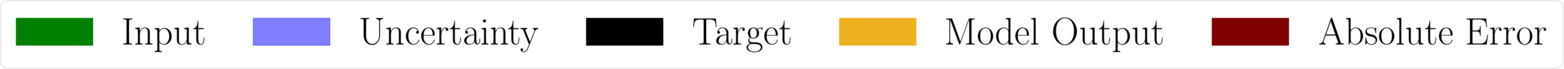}}
    \end{tabular}
    \caption{\textbf{\textsc{1DDeconv} Task Results.} Results for the \textsc{1DDeconv} task on the same sample without noise on the left and with Gaussian noise ($\sigma=0.05$) on the right. The first row displays input and target vector. The figures below show target, network output, together with the uncertainty estimation for the upper graph and the uncertainty estimation plotted against the absolute error in the lower graph for each corresponding method; (I) \textsc{MCDrop}, (II) \textsc{ProbOut}, (III) \textsc{INN}.}
\end{figure*}

\begin{figure}
    \setlength{\fboxsep}{0pt}%
    \setlength{\fboxrule}{1pt}%
    \begin{tabular}{cc}
         Input & Target\\
         \fbox{\includegraphics[width=0.28\textwidth]{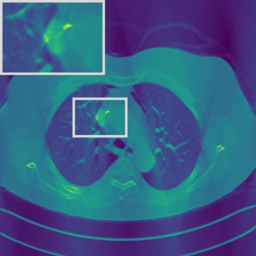}} &
         \fbox{\includegraphics[width=0.28\textwidth]{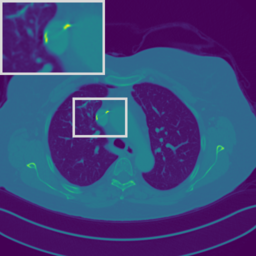}}
    \end{tabular}
    \centering
    \vspace{0.15cm}
    \hspace*{0.05\textwidth}
    \includegraphics[width=0.9466\textwidth]{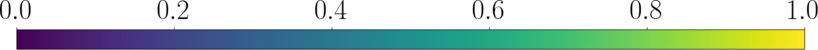}
    
    \begin{tabular}{rccc}
         &Dropout & Probout & Interval \\
         \rotatebox{90}{Prediction}&\fbox{\includegraphics[width=0.28\textwidth]{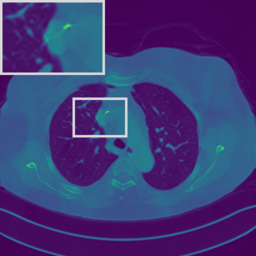}} &
         \fbox{\includegraphics[width=0.28\textwidth]{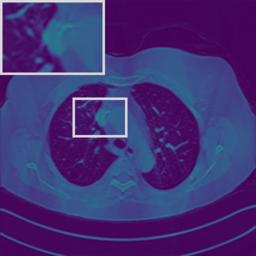}} &
         \fbox{\includegraphics[width=0.28\textwidth]{figures/CTImage2/dropoutimg_output.png}} \\
         \rotatebox{90}{Uncertainty}&\fbox{\includegraphics[width=0.28\textwidth]{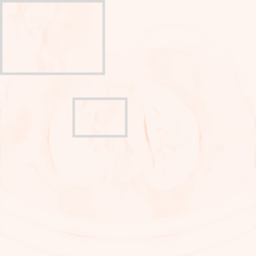}} &
         \fbox{\includegraphics[width=0.28\textwidth]{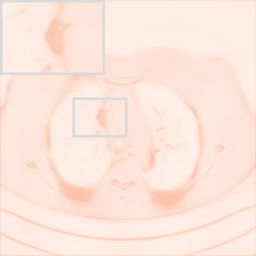}} &
         \fbox{\includegraphics[width=0.28\textwidth]{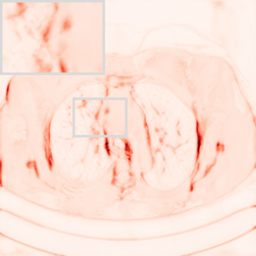}} \\
         \rotatebox{90}{Absolute Error}&\fbox{\includegraphics[width=0.28\textwidth]{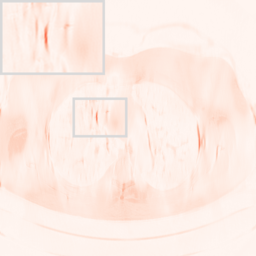}} &
         \fbox{\includegraphics[width=0.28\textwidth]{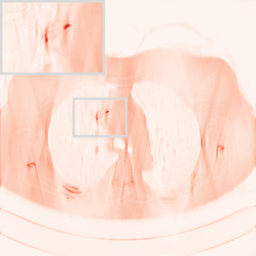}} &
         \fbox{\includegraphics[width=0.28\textwidth]{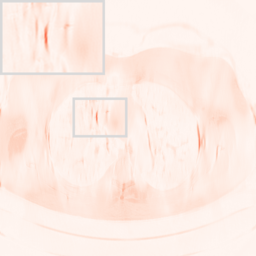}}
    \end{tabular}\\
    
    \vspace*{0.1cm}
    \hspace*{0.05\textwidth}
    \includegraphics[width=0.9466\textwidth]{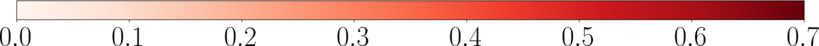}
    
    \caption{\textbf{\textsc{CT} Task Results.} Top row displays input (left) and corresponding target (right). Corresponding predictions (second row), uncertainty scores as standard deviation (\DropShort\ and \ProbShort) and interval size (\IntervalShort) (third row) as well as absolute errors (fourth row) are displayed below for each of the uncertainty methods.}
\end{figure}
\begin{figure}
    \setlength{\fboxsep}{0pt}%
    \setlength{\fboxrule}{1pt}%
    \begin{tabular}{cc}
         Input & Target\\
         \fbox{\includegraphics[width=0.28\textwidth]{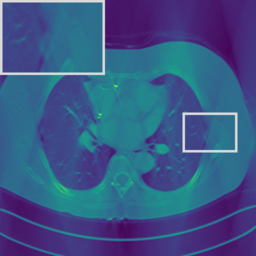}} &
         \fbox{\includegraphics[width=0.28\textwidth]{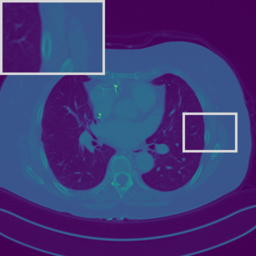}}
    \end{tabular}
    \centering
    \vspace{0.15cm}
    \hspace*{0.05\textwidth}
    \includegraphics[width=0.9466\textwidth]{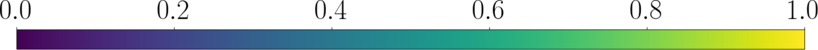}
    
    \begin{tabular}{rccc}
         &Dropout & Probout & Interval \\
         \rotatebox{90}{Prediction}&\fbox{\includegraphics[width=0.28\textwidth]{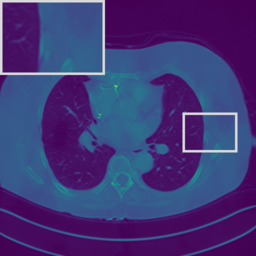}} &
         \fbox{\includegraphics[width=0.28\textwidth]{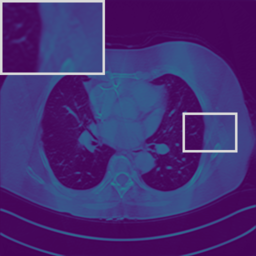}} &
         \fbox{\includegraphics[width=0.28\textwidth]{figures/CTImage3/dropoutimg_output.png}} \\
         \rotatebox{90}{Uncertainty}&\fbox{\includegraphics[width=0.28\textwidth]{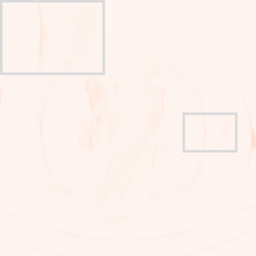}} &
         \fbox{\includegraphics[width=0.28\textwidth]{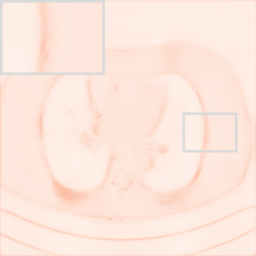}} &
         \fbox{\includegraphics[width=0.28\textwidth]{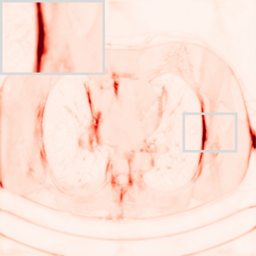}} \\
         \rotatebox{90}{Absolute Error}&\fbox{\includegraphics[width=0.28\textwidth]{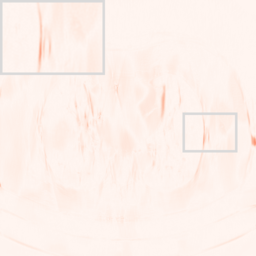}} &
         \fbox{\includegraphics[width=0.28\textwidth]{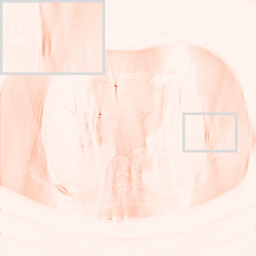}} &
         \fbox{\includegraphics[width=0.28\textwidth]{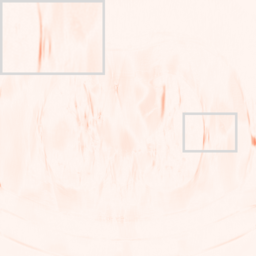}}
    \end{tabular}\\
    
    \vspace*{0.1cm}
    \hspace*{0.05\textwidth}
    \includegraphics[width=0.9466\textwidth]{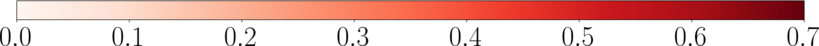}
    
    \caption{\textbf{\textsc{CT} Task Results.} Top row displays input (left) and corresponding target (right). Corresponding predictions (second row), uncertainty scores as standard deviation (\DropShort\ and \ProbShort) and interval size (\IntervalShort) (third row) as well as absolute errors (fourth row) are displayed below for each of the uncertainty methods.}
\end{figure}

\end{appendices}

\end{document}